\pdfoutput=1

\documentclass[10pt,twocolumn,letterpaper]{article}

\usepackage{iccv}
\usepackage{times}
\usepackage{epsfig}
\usepackage{graphicx}
\usepackage{amsmath}
\usepackage{amssymb}
\usepackage{subfigure}
\usepackage{multirow}
\usepackage{caption}
\usepackage{booktabs}       %
\usepackage{standalone}
\usepackage{xspace}
\usepackage{enumitem}
\usepackage{url}
\usepackage{array}
\usepackage{fancyhdr}
\usepackage[pagebackref=true,breaklinks=true,letterpaper=true,colorlinks,bookmarks=false]{hyperref} %
\iccvfinalcopy %

 \makeatletter
 \DeclareRobustCommand\onedot{\futurelet\@let@token\@onedot}
 \def\@onedot{\ifx\@let@token.\else.\null\fi\xspace}
 \def\eg{e.g\onedot} 
 \def\ie{i.e\onedot}

 \def\wrt{w.r.t\onedot}

 \makeatother

\DeclareRobustCommand{\Figref}[1]{Figure~\ref{#1}}

\DeclareRobustCommand{\Secref}[1]{Section~\ref{#1}}
\DeclareRobustCommand{\secref}[1]{Section~\ref{#1}}

\DeclareRobustCommand{\Tableref}[1]{Table~\ref{#1}}

\DeclareRobustCommand{\Eqnref}[1]{Equation~(\ref{#1})}

\newcommand{\myparagraph}[1]{\vspace{2pt}\noindent{\bf #1}}

\iccvfinalcopy %

\def\iccvPaperID{1731} %

\newcommand{\red}[1]{\leavevmode\begin{color}{red}#1\end{color}}

\newcommand{\authSpace}{\quad}
\newcommand*{\mybullet}{\textcolor[rgb]{0.3,0.1,0.5}{\textbullet\,}}
\newcolumntype{x}[1]{>{\raggedright\arraybackslash\hspace{0pt}}p{#1}}

\ificcvfinal\fi %
\begin{document}

\title{Speaking the Same Language:\\Matching Machine to Human Captions by Adversarial Training}

\author{
    Rakshith Shetty$^{1}$ \authSpace Marcus Rohrbach$^{2,3}$ \authSpace Lisa Anne Hendricks$^{2}$\\[2mm]
Mario Fritz$^{1}$  \authSpace Bernt Schiele$^{1}$\\[4mm]
$^{1}$Max Planck Institute for Informatics, Saarland Informatics Campus, Saarbr{\"u}cken, Germany\\[1mm]
$^{2}$UC Berkeley EECS,  CA, United States \ \ \ \ \ \ \ \ \ \ \ \  $^{3}$Facebook AI Research\\[1mm]
}

\maketitle
\thispagestyle{fancy}

\begin{abstract}
While strong progress has been made in image captioning recently, machine and human captions are still quite distinct. This is primarily due to the deficiencies in the generated word distribution, vocabulary size, and strong bias in the generators towards frequent captions. Furthermore, humans -- rightfully so -- generate multiple, diverse captions, due to the inherent ambiguity in the captioning task which is not explicitly considered in today's systems.

To address these challenges, we change the training objective of the caption generator from reproducing ground-truth captions to generating a set of captions that is indistinguishable from human written captions. Instead of handcrafting such a learning target, we employ adversarial training in combination with an approximate Gumbel sampler to implicitly match the generated distribution to the human one.
While our method achieves comparable performance to the state-of-the-art in terms of the correctness of the captions, we generate a set of diverse captions that are significantly less biased and better match the global uni-, bi- and tri-gram distributions of the human captions.

\end{abstract}

\begin{figure}[htp]
\begin{center}
\centering
\footnotesize
\begin{tabular}{p{3.6cm}p{3.6cm}}
  \includegraphics[width=0.40\columnwidth, height=0.25\columnwidth]{images/COCO_val2014_000000225299.jpg}
  &\includegraphics[width=0.40\columnwidth, height=0.25\columnwidth]{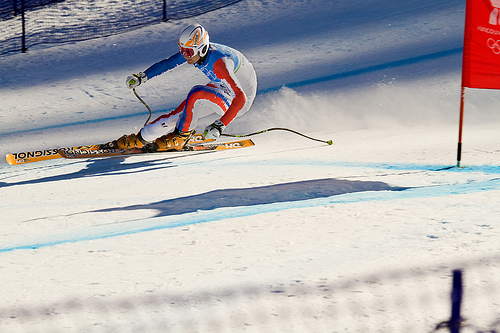}
  \\
  \textbf{Ours}: a person on skis jumping over a ramp& \textbf{Ours}: a skier is making a turn on a course\\
  \includegraphics[width=0.40\columnwidth, height=0.25\columnwidth]{images/COCO_val2014_000000038210.jpg}
  &\includegraphics[width=0.40\columnwidth, height=0.25\columnwidth]{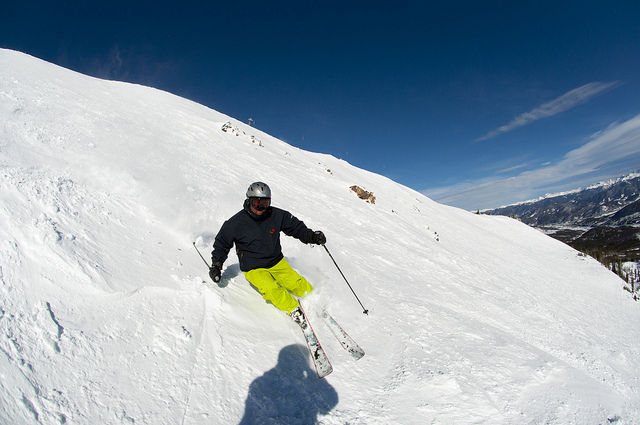}\\
  \textbf{Ours}: a cross country skier makes his way through the snow &\textbf{Ours}: a skier is headed down a steep slope \\
\cmidrule(lr){1-2}
\multicolumn{2}{c}{\textbf{Baseline}: a man riding skis down a snow covered slope}\\
\cmidrule(lr){1-2}
\end{tabular}
  \caption{Four images from the test set, all related to skiing, shown with captions from our adversarial model and a baseline. Baseline model describes all four images with one generic caption, whereas our model produces diverse and more image specific captions.}
  \label{fig:teaser-multiimage}
  \vspace{-2em}
\end{center}
\end{figure}

\begin{figure}[htp]
\begin{center}
\centering
\footnotesize
\begin{tabular}{p{3mm}p{3.5cm}p{3.5cm}}
  &\includegraphics[width=0.4\columnwidth, height=0.35\columnwidth]{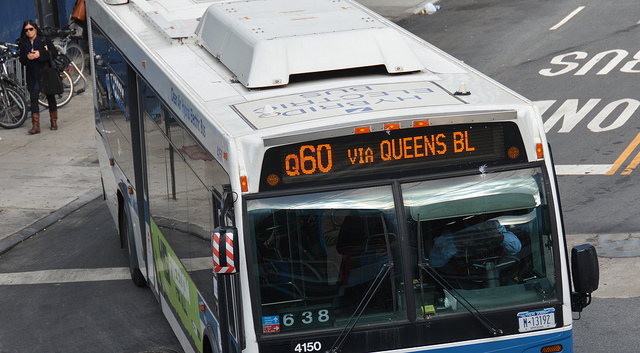}
  & \includegraphics[width=0.4\columnwidth, height=0.35\columnwidth]{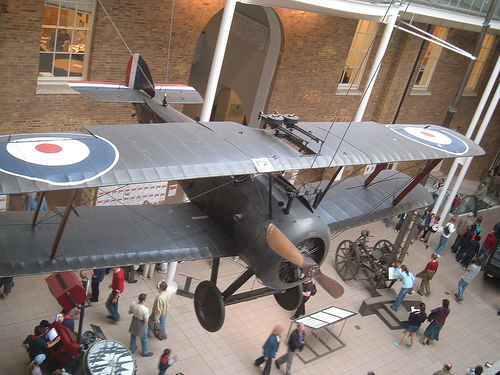}
  \\
  \cmidrule(lr){2-2}  \cmidrule(lr){3-3}
  \multirow{2}{*}{Ours} & a bus that has pulled into the side of the street &\red{a group of} people standing outside \red{in a} old museum\\
  & a bus is parked at the side of the road& an airplane show where people stand around\\
  & a white bus is parked near a curb with people walking by& a line of planes parked at an airport show\\
  \cmidrule(lr){2-2}  \cmidrule(lr){3-3}
  \multirow{2}{*}{\parbox{1mm}{Base \\ line}} & \mybullet a bus is parked \red{on the} side of the road & \red{a group of} people standing around a plane\\
  & \mybullet a bus that is parked \red{in the} street & \red{a group of} people standing around a plane\\
  & a bus is parked \red{in the} street \red{next to a} bus& \red{a group of} people standing around a plane\\
\cmidrule(lr){2-3}
\end{tabular}
  \caption{Two examples comparing multiple captions generated by our adversarial model and the baseline. Bi-grams which are top-20 frequent bi-grams in the training set are marked in red (e.g., ``a group'' and ``group of''). Captions which are replicas from training set are marked with \mybullet.
}
  \label{fig:teaser-multicap}
  \vspace{-2em}
\end{center}
\end{figure}

\section{Introduction}
Image captioning systems have a variety of applications ranging from media retrieval and tagging to assistance for the visually impaired. In particular, models which combine state-of-the-art image representations based on deep convolutional networks and deep recurrent language models have led to ever increasing performance on evaluation metrics such as CIDEr~\cite{Vedantam_2015_CVPR} and METEOR~\cite{lavie2014meteor} as can be seen \eg on the COCO image Caption challenge leaderboard \cite{cocoleaderboard}. 

Despite these advances, it is often easy for humans to differentiate between machine and human captions -- particularly when observing multiple captions for a single image. As we analyze in this paper, this is likely due to artifacts and deficiencies in the statistics of the generated captions, which is more apparent when observing multiple samples. Specifically, we observe that state-of-the-art systems frequently ``reveal themselves'' by generating a different word distribution and using smaller vocabulary. Further scrutiny reveals that generalization from the training set is still challenging and generation is biased to frequent fragments and captions.

Also, today's systems are evaluated to produce a single caption. Yet, multiple potentially distinct captions are typically correct for a single image -- a property that is reflected in human ground-truth. This diversity is not equally reproduced by state-of-the-art caption generators \cite{vijayakumar2016diverse, li16naacl}.

Therefore, our goal is to make image captions less distinguishable from human ones -- similar in the spirit to a Turing Test. We also embrace the ambiguity of the task and extend our investigation to predicting sets of captions for a single image and evaluating their quality, particularly in terms of the diversity in the generated set. In contrast, popular approaches to image captioning are trained with an objective to reproduce the captions as provided by the ground-truth.

Instead of relying on handcrafting loss-functions to achieve our goal, we propose an adversarial training mechanism for image captioning. For this we build on Generative Adversarial Networks~(GANs)~\cite{goodfellow2014generative}, which have been successfully used to generate mainly continuous data distributions such as images~\cite{denton15nips,radford16iclr}, although exceptions exist \cite{luc2016semantic}. In contrast to images, captions are discrete, which poses a challenge when trying to backpropagate through the generation step. To overcome this obstacle, we use a Gumbel sampler \cite{jang2016categorical,maddison2016concrete} that allows for end-to-end training.

We address the problem of caption set generation for images and discuss metrics to measure the caption diversity and compare it to human ground-truth. We contribute a novel solution to this problem using an adversarial formulation. The evaluation of our model shows that accuracy of generated captions is on par to the state-of-the-art, but we greatly increase the diversity of the caption sets and better match the ground-truth statistics in several measures. Qualitatively, our model produces more diverse captions across images containing similar content (\Figref{fig:teaser-multiimage}) and when sampling multiple captions for an image (see supplementary)\footnote{\url{https://goo.gl/3yRVnq} }.%

\section{Related Work} 
\myparagraph{Image Description.} Early captioning models rely on first recognizing visual elements, such as objects, attributes, and activities, and then generating a sentence using language models such as a template model \cite{farhadi10eccv}, n-gram model \cite{kulkarni2013babytalk}, or statistical machine translation \cite{rohrbach13iccv}.  
Advances in deep learning have led to end-to-end trainable models that combine deep convolutional networks to extract visual features and recurrent networks to generate sentences \cite{donahue15cvpr, vinyals2015show, karpathy15cvpr}.

Though modern description models are capable of producing coherent sentences which accurately describe an image, they tend to produce generic sentences which are replicated from the train set \cite{devlin15acl}.
Furthermore, an image can correspond to many valid descriptions.
However, at test time, sentences generated with methods such as beam search are generally very similar.
\cite{vijayakumar2016diverse,li16naacl} focus on increasing sentence diversity by integrating a diversity promoting heuristic into beam search.
\cite{wang16ijai} attempts to increase the diversity in caption generation by training an ensemble of caption generators each specializing in different portions of the training set.
In contrast, we focus on improving diversity of generated captions using a single model. Our method achieves this by learning a corresponding model using a different training loss as opposed to after training has completed.
We note that generating diverse sentences is also a challenge in visual question generation, see concurrent work~\cite{jain2017creativity}, and in language-only dialogue generation studied in the linguistic community, see \eg \cite{li16naacl, li2017adversarial}.

When training recurrent description models, the most common method is to predict a word $w_t$ conditioned on an image and all previous \textit{ground truth} words.
At test time, each word is predicted conditioned on an image and previously \textit{predicted} words.
Consequently, at test time predicted words may be conditioned on words that were incorrectly predicted by the model.
By only training on ground truth words, the model suffers from \emph{exposure bias}~\cite{ranzato2015sequence} and cannot effectively learn to recover when it predicts an incorrect word during training. 
To avoid this, \cite{bengio2015scheduled} proposes a scheduled sampling training scheme which begins by training with ground truth words, but then slowly conditions generated words on words previously produced by the model. However, \cite{Huszar2015HowT} shows that the scheduled sampling algorithm is inconsistent and the optimal solution under this objective does not converge to the true data distribution.
Taking a different direction, \cite{ranzato2015sequence} proposes to address the exposure bias by gradually mixing a sequence level loss~(BLEU score) using REINFORCE rule with the standard maximum likelihood training. Several other works have followed this up with using reinforcement learning based approaches to directly optimize the evaluation metrics like BLEU, METEOR and CIDER~\cite{rennie2016self, liu2016optimization}. However, optimizing the evaluation metrics does not directly address the diversity of the generated captions. Since all current evaluation metrics use n-gram matching to score the captions, captions using more frequent n-grams are likely to achieve better scores than ones using rarer and more diverse n-grams. 

In this work, we formulate our caption generator as a generative adversarial network. We design a discriminator that explicitly encourages generated captions to be diverse and indistinguishable from human captions. The generator is trained with an adversarial loss with this discriminator. Consequently, our model generates captions that better reflect the way humans describe images while maintaining similar correctness as determined by a human evaluation.

\myparagraph{Generative Adversarial Networks.}
The Generative Adversarial Networks (GANs) \cite{goodfellow2014generative} framework learns generative models without explicitly defining a loss from a target distribution. Instead, GANs learn a generator using a loss from a discriminator which tries to differentiate real and generated samples, where the generated samples come from the generator.
When training to generate real images, GANs have shown encouraging results~\cite{denton15nips,radford16iclr}. In all these works the target distribution is continuous. In contrast our target, a sequence of words, is discrete.  Applying GANs to discrete sequences is challenging as it is unclear how to best back-propagate the loss through the sampling mechanism.

A few works have looked at generating discrete distributions using GANs. \cite{luc2016semantic} aim to generate a semantic image segmentation with discrete semantic labels at each pixel.
 \cite{yu2016seqgan} uses REINFORCE trick to train an unconditional text generator using the GAN framework but diversity of the generated text is not considered. 

Most similar to our work are concurrent works which use GANs for dialogue generation \cite{li2017adversarial} and image caption generation \cite{dai2017towards}.
While \cite{li2017adversarial,yu2016seqgan,dai2017towards} rely on the reinforcement rule \cite{williams92Reinforce} to handle backpropagation through the discrete samples, we use the Gumbel Softmax \cite{jang2016categorical}. See \secref{sec:model:generator} for further discussion. \cite{li2017adversarial} aims to generate a diverse dialogue of multiple sentences while we aim to produce diverse sentences for a single image.
Additionally, \cite{li2017adversarial} uses both the adversarial and the maximum likelihood loss in each step of generator training. We however train the generator with only adversarial loss after pre-training. 
Concurrent work \cite{dai2017towards} also applies GANs to diversify generated image captions. Apart from using the gumbel softmax as discussed above, our work differs from~\cite{dai2017towards} in the discriminator design and quantitative evaluation of the generator diversity.

\section{Adversarial Caption Generator}
The image captioning task can be formulated as follows: given an input image
$x$ the generator $G$ produces a caption, $G(x) = [w_0, \dots, w_{n-1}]$,
describing the contents of the image. There is an inherent ambiguity in the task, with multiple possible correct captions for an image, which is also reflected in diverse captions written by human annotators ~(we quantify this in~\Tableref{tab:diversityStat}).
However, most image captioning architectures ignore this diversity during training. The standard approach to model $G(x)$ is to use a recurrent language model conditioned on the input image $x$~\cite{donahue15cvpr,vinyals2015show}, and train it using a maximum likelihood~(ML) loss considering every image--caption pair as an independent sample. This ignores the diversity in the human captions and results in models that tend to produce generic and commonly occurring captions from the training set, as we will show in~\secref{sec:result:stats}. 

We propose to address this by explicitly training the generator $G$ to produce multiple diverse captions for an input image using the adversarial framework~\cite{goodfellow2014generative}. In adversarial frameworks, a generative model is trained by pairing it with adversarial discriminator which tries to distinguish the generated samples from true data samples. The generator is trained with the objective to fool the discriminator, which is optimal when $G$ exactly matches the data distribution. This is well-suited for our goal because, with an appropriate discriminator network we could coax the generator to capture the diversity in the human written captions, without having to explicitly design a loss function for it.

To enable adversarial training, we introduce a second network, $D(x,s)$, which takes as input an image $x$ and a caption set $S_p = \{s_1, \dots, s_p\}$ and classifies it as either real or fake. Providing a set of captions per image as input to the discriminator allows it to factor in the diversity in the caption set during the classification. The discriminator can penalize the generator for producing very similar or repeated captions and thus encourage the diversity in the generator.

Specifically, the discriminator is trained to classify the captions drawn from the reference captions set, $R(x)= \{r_0,\cdots,r_{k-1}\}$, as real while classifying the captions produced by the generator, $G(x)$, as fake. The generator $G$ can now be trained using an adversarial objective, i.e. $G$ is trained to fool the discriminator to classify $G(x)$ as real.

\subsection{Caption generator}
\begin{figure}
\centering
    \includegraphics[width=0.85\columnwidth]{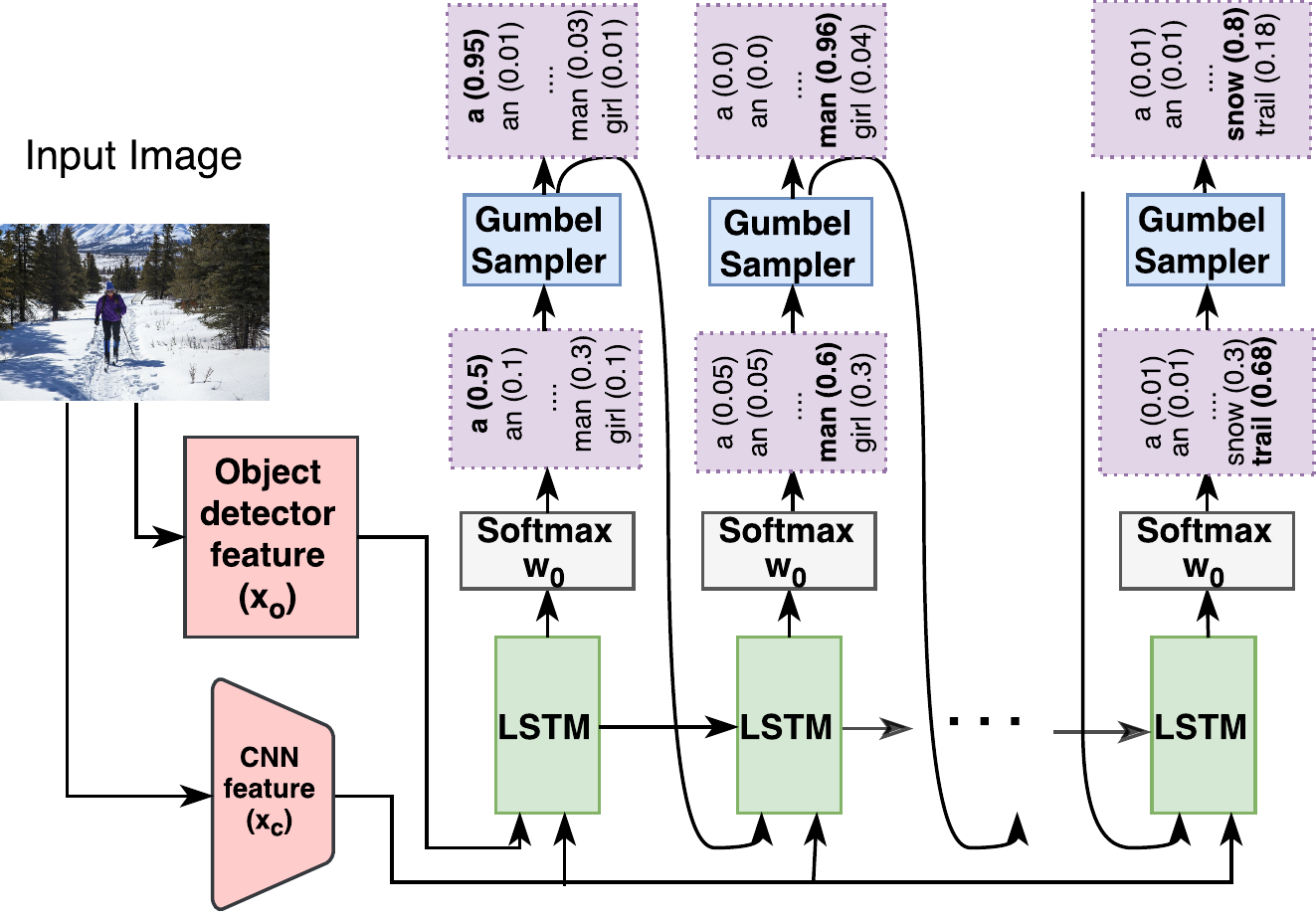}\hfill
   \caption{Caption generator model. Deep visual features are input to an LSTM to generate a sentence. A Gumbel sampler is used to obtain \emph{soft} samples from the softmax distribution, allowing for backpropagation through the samples.}
   \vspace{-0.2cm}
\label{fig:BlkGenerator}
\end{figure}

\label{sec:model:generator}
We use a near state-of-the art caption generator model based on~\cite{ShettyACMMM2016Wrk}.
It uses the standard encoder-decoder framework with two stages: the encoder model which extracts feature vectors from the input image and the decoder which translates these features into a word sequence.

\myparagraph{Image features.} Images are encoded as activations from a pre-trained convolutional neural network~(CNN). Captioning models also benefit from augmenting the CNN features with explicit object detection features~\cite{ShettyACMMM2016Wrk}. Accordingly, we extract a feature vector containing the probability of occurrence of an object and provide it as input to the generator.

\myparagraph{Language Model.} Our decoder shown in \Figref{fig:BlkGenerator}, is adopted from a Long-Short Term Memory~(LSTM) based language model architecture presented
in~\cite{ShettyACMMM2016Wrk} for image captioning. It consists of a
three-layered LSTM network with residual connections between the layers. The
LSTM network takes two features as input. First is the object detection feature, $x_{o}$, which is input to the LSTM at only $0th$ time step and shares the input matrix with the word vectors. Second is the global image CNN feature, $x_{c}$, and is input to the LSTM at all time-steps through its own input matrix.

The softmax layer at the output of the generator produces a probability
distribution over the vocabulary at each step.
\begin{align}
    y_t &= \text{LSTM}(w_{t-1}, x_c, y_{t-1}, c_{t-1})\\
    \label{eqn:softmax}
    p(w_t|w_{t-1},x) &= \text{softmax}\left[\beta W_d*y_t\right],
\end{align}
where $c_t$ is the LSTM cell state at time $t$ and $\beta$ is a scalar parameter which controls the peakyness of the distribution. Parameter $\beta$ allows us to control how large a hypothesis space the generator explores during adversarial training.
An additional uniform random noise vector $z$, is input to the LSTM in adversarial training to allow the generator to use the noise to produce diversity.

\myparagraph{Discreteness Problem.} To produce captions from the generator we could simply sample from this distribution $p(w_t|w_{t-1},x)$, recursively feeding back the previously sampled word at each step, until we sample the END token. One can generate multiple sentences by sampling and pick the sentence with the highest probability as done in \cite{donahue16pami}. Alternatively we could also use greedy search approaches like beam-search. 
However, directly providing these discrete samples as input to the
discriminator does not allow for backpropagation through them as they are discontinuous. %
Alternatives to overcome this are the reinforce rule/trick \cite{williams92Reinforce}, using the softmax distribution, or using the  Gumbel-Softmax approximation \cite{jang2016categorical,maddison2016concrete}.

   Using policy gradient algorithms with the reinforce rule/trick \cite{williams92Reinforce} allows estimation of gradients through discrete samples~\cite{hendricks16eccv,andreas16emnlp, yu2016seqgan,li2017adversarial}. However, learning using reinforce trick can be unstable due to high variance~\cite{sutton1998reinforcement} and some mechanisms to make learning more stable, like estimating the action-value for intermediate states by generating multiple possible sentence completions~(e.g used in~\cite{yu2016seqgan,dai2017towards}), can be computationally intensive.
  
  Another option is to input the softmax distribution to the discriminator instead of samples. We experimented with this, but found that the discriminator easily distinguishes between the softmax distribution produced by the generator and the sharp reference samples, and the GAN training fails.
  
 The last option, which we rely on in this work, it to use a continuous relaxation of the samples encoded as one-hot vectors using the Gumbel-Softmax approximation proposed in \cite{jang2016categorical} and \cite{maddison2016concrete}. This continuous relaxation combined with the re-parametrization of the sampling process allows backpropagation through samples from a categorical distribution. The main benefit of this approach is that it plugs into the model as a differentiable node and does not need any additional steps to estimate the gradients. Whereas most previous methods to applying GAN to discrete output generators use policy gradient algorithms, we show that Gumbel-Softmax approximation can also be used successfully in this setting. An empirical comparison between the two approaches can be found in~\cite{jang2016categorical}.

The Gumbel-Softmax approximation consists of two steps. First Gumbel-Max trick is used to re-parametrize sampling from a categorical
distribution. Given a random variable $r$ drawn from a categorical
distribution parametrized by $\Theta = {\theta_0, \cdots, \theta_{v-1}}$, $r$ can be expressed as:
\begin{equation}
    r = \text{one\_hot}\left[\text{arg}\max_i\left(g_i +\log\theta_i\right)\right],
\label{eqn:gumbelmax}
\end{equation}
where $g_i$'s are i.i.d. random variables from the standard gumbel distribution.
Next the $\text{argmax}$ in \Eqnref{eqn:gumbelmax} is replaced with softmax to
obtain a continuous relaxation of the discrete random variable $r$. 
\begin{equation}
    r' = \text{softmax}\left[\frac{g_i +\log\theta_i}{\tau}\right],
\label{eqn:gumbeapprox}
\end{equation}
where $\tau$ is the temperature parameter which controls how close $r'$ is to
$r$, with $r' = r$ when $\tau=0$.

We use straight-through variation of the Gumbel-Softmax
approximation~\cite{jang2016categorical} at the output of our generator to
sample words during the adversarial training. In the straight-through variation,
sample $r$ is used in the forward path and soft approximation $r'$ is used in
the backward path to allow backpropogation.

\subsection{Discriminator model}

\begin{figure}
\centering
    \includegraphics[width=0.9\columnwidth]{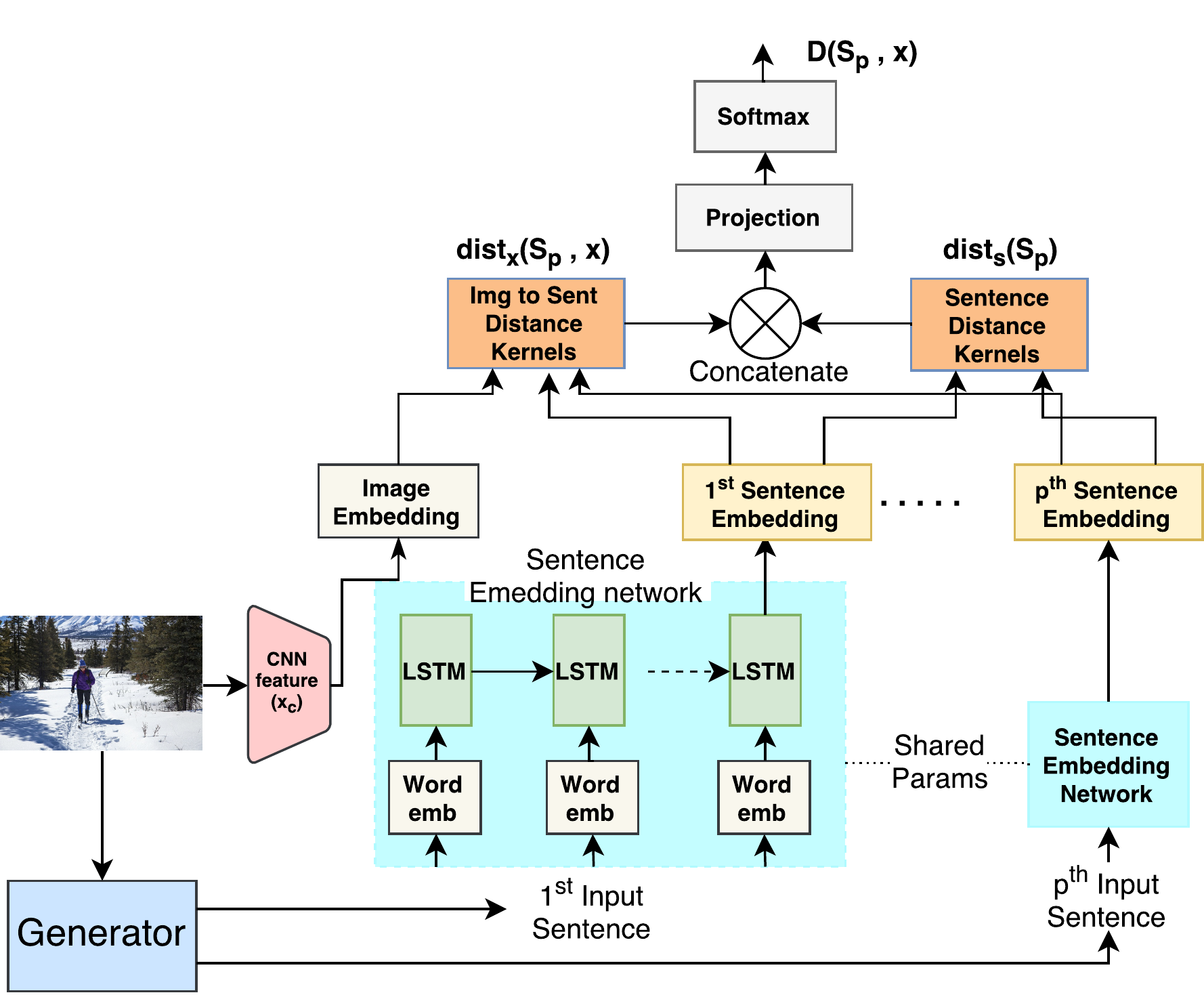}
   \caption{Discriminator Network. Caption set sampled from the generator is used to compute image to sentence $(dist_x(S_p, x))$ and sentence-to-sentence $(dist_s(S_p))$ distances. They are used to score the set as real/fake.}
\label{fig:BlkDiscriminator}
\end{figure}

The discriminator network, $D$ takes an image $x$, represented using CNN feature $x_c$, and a set  of captions $S_p=\{s_1,\dots, s_{p}\}$  as input  and classifies $S_p$ as either real or fake.
Ideally, we want $D$ to base this decision on two criteria: a)
do $s_i\in S_p$ describe the image correctly ? b) is the set $S_p$ is
diverse enough to match the diversity in human captions ? 

To enable this, we use two separate distance measuring kernels in our
discriminator network as shown in \Figref{fig:BlkDiscriminator}. The first
kernel computes the distances between the image $x$ and each sentence in $S_p$.
The second kernel computes the distances between the sentences in $S_p$. The
architecture of these distance measuring kernels is based on the minibatch
discriminator presented in~\cite{salimans2016improved}. However, unlike~\cite{salimans2016improved}, we
only compute distances between captions corresponding to the same image and not over the entire minibatch.

Input captions are encoded into a fixed size sentence embedding vector
using an LSTM encoder to obtain vectors $f(s_i) \in \mathbb{R}^M$. The image feature, $x_c$, is also embedded into a smaller image embedding vector $f(x_c)\in \mathbb{R}^M$. The
distances between $f(s_i), i\in\{1,\dots,p\}$ are computed as 
\begin{align}
    K_i &= T_s \cdot f(s_i)\\ 
    c_l(s_i, s_j) &= exp\left(-\|K_{i,l} - K_{j,l}\|_{L_1}\right)\\
    d_l(s_i) &= \sum_{j=1}^{p} c_l(s_i,s_j) \\ 
    dist_s(S_p) &= \left[d_1(s_1),..., d_O(s_1),..., d_O(s_p)\right]\in\mathbb{R}^{p\times O}
\end{align}
where $T_s$ is a $M\times N\times O$ dimensional tensor and $O$ is the number of different $M\times N$ distance kernels to use.

Distances between $f(s_i), i\in{1,\dots,p}$ and $f(x_c)$ are obtained with similar procedure as above, but using a different tensor $T_x$ of dimensions $M\times N\times O$ to yield $dist_x(S_p, x) \in \mathbb{R}^{p\times O}$. These two distance vectors capture the two aspects we want our discriminator to focus on. $dist_x(S_p,x)$ captures how well $S_p$ matches the image $x$ and $dist_s(S_p)$ captures the diversity in $S_p$. The two distance vectors are
concatenated and multiplied with a output matrix followed
by softmax to yield the discriminator output probability, $D(S_p,x)$, for $S_p$ to
be drawn from reference captions.

\subsection{Adversarial Training}
In adversarial training both the generator and the discriminator are trained
alternatively for $n_g$ and  $n_d$ steps respectively. The discriminator tries to
 classify $S^r_p\in R(x)$ as real and $S^g_p \in G(x)$ as fake. In
addition to this, we found it important to also train the discriminator to
classify  few reference captions drawn from a random image as fake, \ie
$S^f_p\in R(y), y\neq x$.  This forces the discriminator to learn to match images
and captions, and not just rely on diversity statistics of the caption set.
The complete loss function of the discriminator is defined by
\begin{multline}
    L(D) = -\log\left(D(S^r_p,x)\right) - \log\left(1-D(S^g_p,x)\right) \\-
    \log\left(1-D(S^f_p,x)\right)
\end{multline}

The training objective of the generator is to fool the discriminator into
classifying $S^g_p\in G(x)$ as real. We found helpful to additionally use the
feature matching loss~\cite{salimans2016improved}. This loss trains the
generator to match activations induced by the generated and true data at some
intermediate layer of the discriminator. In our case we use an $l_2$ loss to match the expected value of distance vectors $dist_s(S_p)$ and $dist_x(S_p, x)$ between real and generated data. The
generator loss function is given by
\begin{equation}
\begin{split}
    L(G) = - &\log\left(D(S^g_p,x)\right)\\
    + &\|\mathbb{E}\left[dist_s(S^g_p)\right] -
    \mathbb{E}\left[dist_s(S^r_p)\right]\|_2\\
    + &\|\mathbb{E}\left[dist_x(S^g_p,x)\right] -
    \mathbb{E}\left[dist_x(S^r_p,x)\right]\|_2,
\end{split}
\end{equation}
where the expectation is over a training mini-batch.

\section{Experimental Setup}
We conduct all our experiments on the MS-COCO dataset~\cite{Chen2015}. The training set consists of ~83k images with five human captions each. We use the publicly available test split of 5000 images~\cite{karpathy15cvpr} for all our experiments. ~\secref{sec:result:ablation} uses a validation split of 5000 images. 

For image feature extraction, we use activations from \emph{res5c} layer of the 152-layered \emph{ResNet}~\cite{he16cvpr} convolutional neural network~(CNN)
pre-trained on  ImageNet. The input images are scaled to
$448\times448$ dimensions for \emph{ResNet} feature extraction. Additionally we use features from the VGG network~\cite{simonyan15iclr} in our ablation study in \secref{sec:result:ablation}. Following~\cite{ShettyACMMM2016Wrk}, we additionally extract 80-dimensional object detection features using a Faster Region-Based Convolutional Neural Network~(RCNN)~\cite{ren15fasterrcnn} trained on the 80 object categories in the COCO dataset. The CNN features are input to both the generator (at $x_p$) and the discriminator. Object detection features are input only to the generator at the $x_i$ input and is used in all the generator models reported here.

\subsection{Insights in Training the GAN}
As is well known~\cite{arjovsky2017towards}, we found GAN training to be sensitive to hyper-parameters. Here we discuss some settings
which helped stabilize the training of our models.

We found it necessary to pre-train the generator using standard maximum
likelihood training. Without pre-training, the generator gets stuck producing
incoherent sentences made of random word sequences. We also found pre-training
the discriminator on classifying correct image-caption pairs against random
image-caption pairs helpful to achieve stable GAN training.
We train the discriminator for 5 iterations for every generator update.
We also periodically monitor the classification accuracy of the
discriminator and train it further if it drops below 75\%. This prevents the generator from updating using a bad discriminator.

Without the feature matching term in the generator loss, the GAN training
was found to be unstable and needed additional maximum likelihood update to stabilize it. This was also reported
in~\cite{li2017adversarial}.  However with the feature matching loss, training is stable and the ML update is not needed.

A good range of values for the Gumbel temperature was found to be $(0.1, 0.8)$.
Beyond this range training was unstable, but within this range
the results were not sensitive to it. We use a fixed temperature setting of
$0.5$ in the experiments reported here. The softmax scaling factor, $\beta$ in \eqref{eqn:softmax}, is set to value $3.0$ for training of all the adversarial models reported here. The sampling results are also with $\beta=3.0$.

\section{Results}
We conduct experiments to evaluate our adversarial caption generator w.r.t. two aspects: how human-like  the generated captions are and how accurately they describe the contents of the image. Using diversity statistics and word usage statistics as a proxy for measuring how closely the generated captions mirror the distribution of the human reference captions, we show that the adversarial model is more human-like than the baseline. Using human evaluation and automatic metrics we also show that the captions generated by the adversarial model performs similar to the baseline model in terms of correctness of the caption.

Henceforth, \emph{Base} and \emph{Adv} refer to the baseline and adversarial models, respectively. Suffixes \emph{bs} and \emph{samp} indicate decoding using beamsearch and sampling respectively.

\subsection{Measuring if captions are human-like}
\label{sec:divstatsdef}
\myparagraph{Diversity.}
We analyze $n$-gram usage statistics, compare vocabulary sizes and other diversity
metrics presented below to understand and measure the gaps between human written captions and the automatic methods and show that the adversarial training helps bridge some of these gaps. 

To measure the corpus level diversity of the generated captions we use:
\begin{itemize}[noitemsep,topsep=0pt]
    \item\textit{Vocabulary Size} - number of unique words used in all generated captions 
    \item\textit{\% Novel Sentences} - percentage of generated captions not seen in the training set.
\end{itemize}
To measure diversity in a set of captions, $S_p$, corresponding to a single
image we use:
\begin{itemize}[noitemsep,topsep=0pt]
    \item\textit{Div-1} - ratio of number of unique unigrams in $S_p$ to number of words in $S_p$. Higher is more diverse.
    \item\textit{Div-2} - ratio of number of unique bigrams in $S_p$ to number of words in $S_p$.  Higher is more diverse.
    \item\textit{mBleu} - Bleu score is computed between each caption in $S_p$ against the rest. Mean of these $p$ Bleu scores is the mBleu score. Lower values indicate more diversity.
\end{itemize}
\myparagraph{Correctness.}
Just generating diverse captions is not useful if they do not correctly
describe the content of an image. To measure the correctness of the generated
captions we use two automatic evaluation metrics
Meteor~\cite{lavie2014meteor} and SPICE~\cite{spice2016}. However
since it is known that the automatic metrics do not always correlate very well with
human judgments of the correctness, we also report results from
human evaluations comparing the baseline model to our adversarial model.

\subsection{Comparing caption accuracy}
\Tableref{tab:evalMetrics} presents the comparison of our adversarial model to the baseline model. Both the baseline and the adversarial models use
\emph{ResNet} features. The beamsearch results are with beam size 5 and sampling
results are with taking the best of 5 samples. Here the best caption is obtained by ranking the captions as per probability assigned by the model.

\Tableref{tab:evalMetrics} also shows the metrics from some recent methods from
the image captioning literature. The purpose of this comparison is to illustrate
that we use a strong baseline and that our baseline model is competitive to
recent published work, as seen from the Meteor and Spice metrics. 

Comparing baseline and adversarial models in \Tableref{tab:evalMetrics} the
adversarial model does worse in-terms of Meteor scores and overall spice
metrics. When we look at Spice scores on individual categories shown in \Tableref{tab:evalMetricsSpice} we see that
adversarial models excel at counting relative to the baseline and describing the size of an object correctly.

However, it is well known that automatic metrics do not always correlate with human judgments on correctness of a caption. A primary reason the adversarial models do poorly on automatic metrics is that they produce significantly more unique sentences using a much larger vocabulary and rarer n-grams, as shown in ~\secref{sec:result:stats}. Thus, they are less likely to do well on metrics relying on $n$-gram matches.

To verify this claim, we conduct human evaluations comparing captions from the
baseline and the adversarial model. Human evaluators from Amazon Mechanical Turk are shown an image and a
caption each from the two models and are asked ``Judge which of the two sentences is
a better description of the image (\wrt correctness and relevance)!''. The choices were either of the two sentences or to report that they are the same. Results
from this evaluation are presented in \Tableref{tab:human}. We can see that both adversarial and baseline models perform similarly, with adversarial models doing slightly better. This shows that despite the poor performance in automatic evaluation metrics, the adversarial models produce captions that are similar, or even slightly better, in accuracy to the baseline model.

\begin{table}[tb]
\centering
\small
\begin{tabular}{lcc}
\toprule
\bf Method & \bf Meteor & \bf Spice\\              
 \cmidrule(lr){1-1}\cmidrule(lr){2-2}\cmidrule(lr){3-3}
   ATT-FCN~\cite{you2016image} & 0.243 &-- \\
   MSM~\cite{yao2016boosting}    & 0.251 &-- \\
   KWL~\cite{lu17cvpr}    & 0.266&\bf 0.194 \\
 \cmidrule(lr){1-3}
 {Ours Base-bs }      &\bf0.272 &0.187 \\
 {Ours Base-samp}    & 0.265 & 0.186 \\
 {Ours Adv-bs}   & 0.239 & 0.167 \\
 {Ours Adv-samp} & 0.236 & 0.166 \\
 \bottomrule
\end{tabular}
\caption{Meteor and Spice metrics comparing performance of baseline and adversarial models.}
\label{tab:evalMetrics}
\end{table}
\begin{table}[tb]
\centering
\small
\resizebox{\columnwidth}{!}{%
\begin{tabular}{lcccccc}
\toprule
\multirow{2}{*}{\bf Method }& \multicolumn{6}{c}{\bf Spice}\\
\cmidrule(lr){2-7}
                            & Color & Attribute & Object & Relation& Count & Size \\
 \cmidrule(lr){1-1}\cmidrule(lr){2-7}
 {Base-bs}    &\bf0.101 &\bf0.085 & 0.345 & 0.049 & 0.025 & 0.034 \\
 {Base-samp}       & 0.059 & 0.069 &\bf0.352 &\bf0.052 & 0.032 & 0.033 \\
 {Adv-bs} & 0.079 & 0.082 & 0.318 & 0.034 &\bf 0.080 & 0.052 \\
 {Adv-samp}    & 0.078 & 0.082 & 0.316 & 0.033 & 0.076 &\bf 0.053 \\
 \bottomrule
\end{tabular}
}
\caption{Comparing baseline and adversarial models in different categories of Spice metric.}
\label{tab:evalMetricsSpice}
\end{table}
\begin{table}[tb]
\centering
\small
\begin{tabular}{lcc}
\toprule
\bf Comparison &  Adversarial - Better & Adversarial -  Worse\\ \cmidrule(lr){1-1}\cmidrule(lr){2-3}
Beamsearch & 36.9 &	34.8\\
Sampling & 35.7 & 33.2\\
 \bottomrule
\end{tabular}
\caption{Human evaluation comparing adversarial model vs the baseline model on
    482 random samples. Correctness of captions. With agreement of at least 3 out of 5 judges in
    \%. Humans agreed in 89.2\% and 86.7\% of images in beamsearch and sampling cases respectively.}
\label{tab:human}
\vspace{-1.5em}
\end{table}
\subsection{Comparing vocabulary statistics}
\label{sec:result:stats}
\begin{figure}
\centering
    \includegraphics[width=0.49\columnwidth]{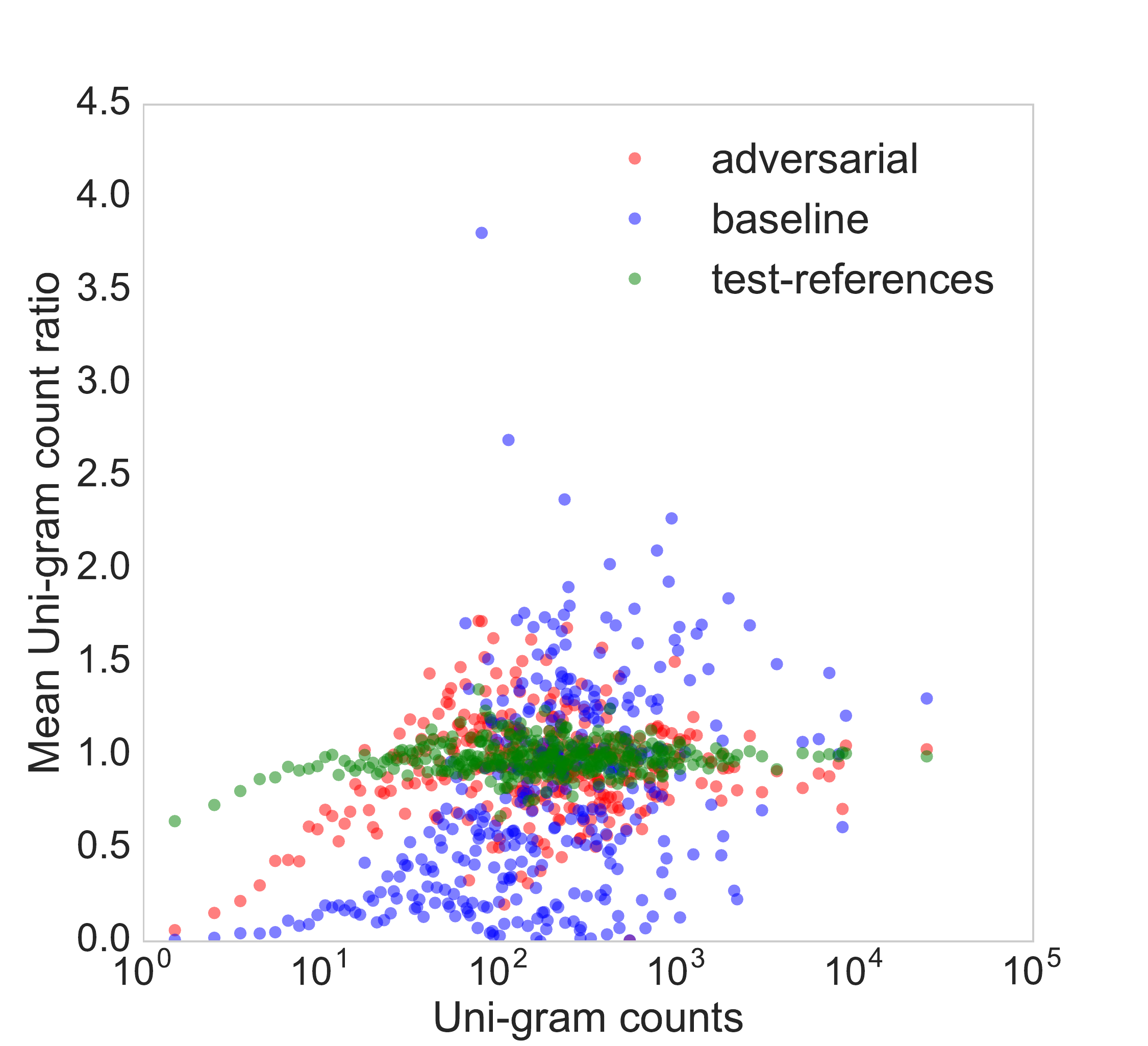}
    \includegraphics[width=0.49\columnwidth]{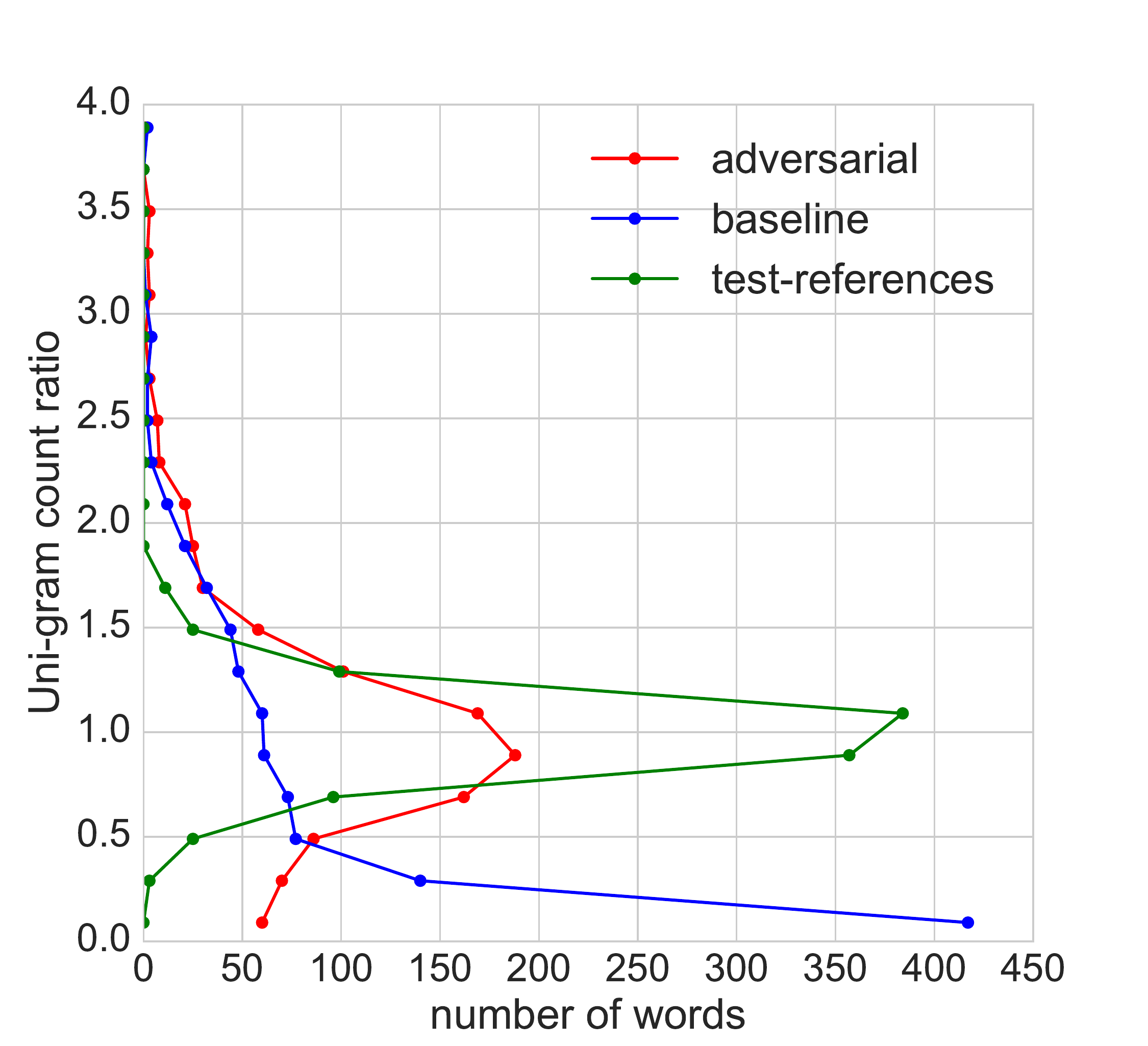}
    \includegraphics[width=0.49\columnwidth]{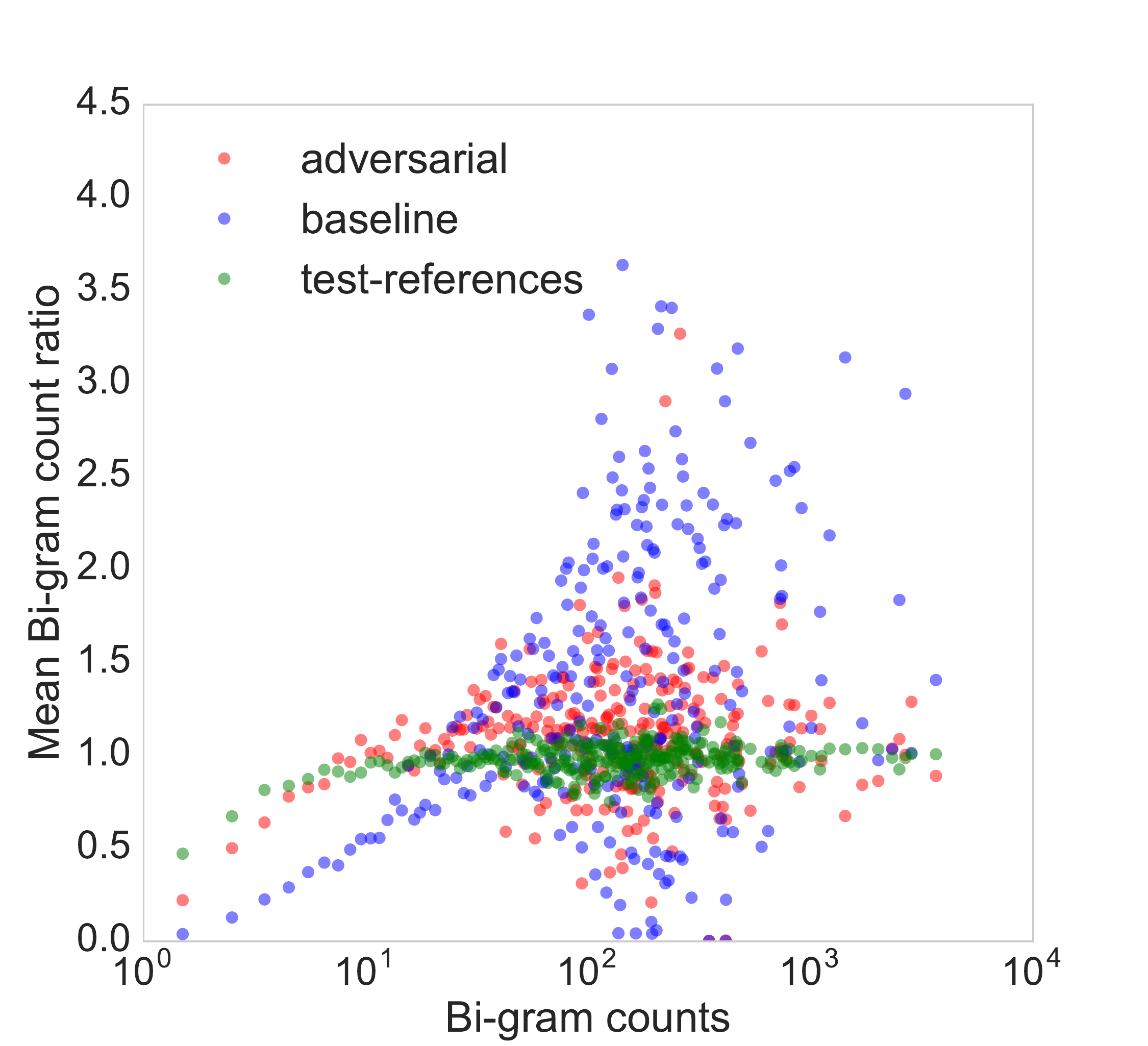}
    \includegraphics[width=0.49\columnwidth]{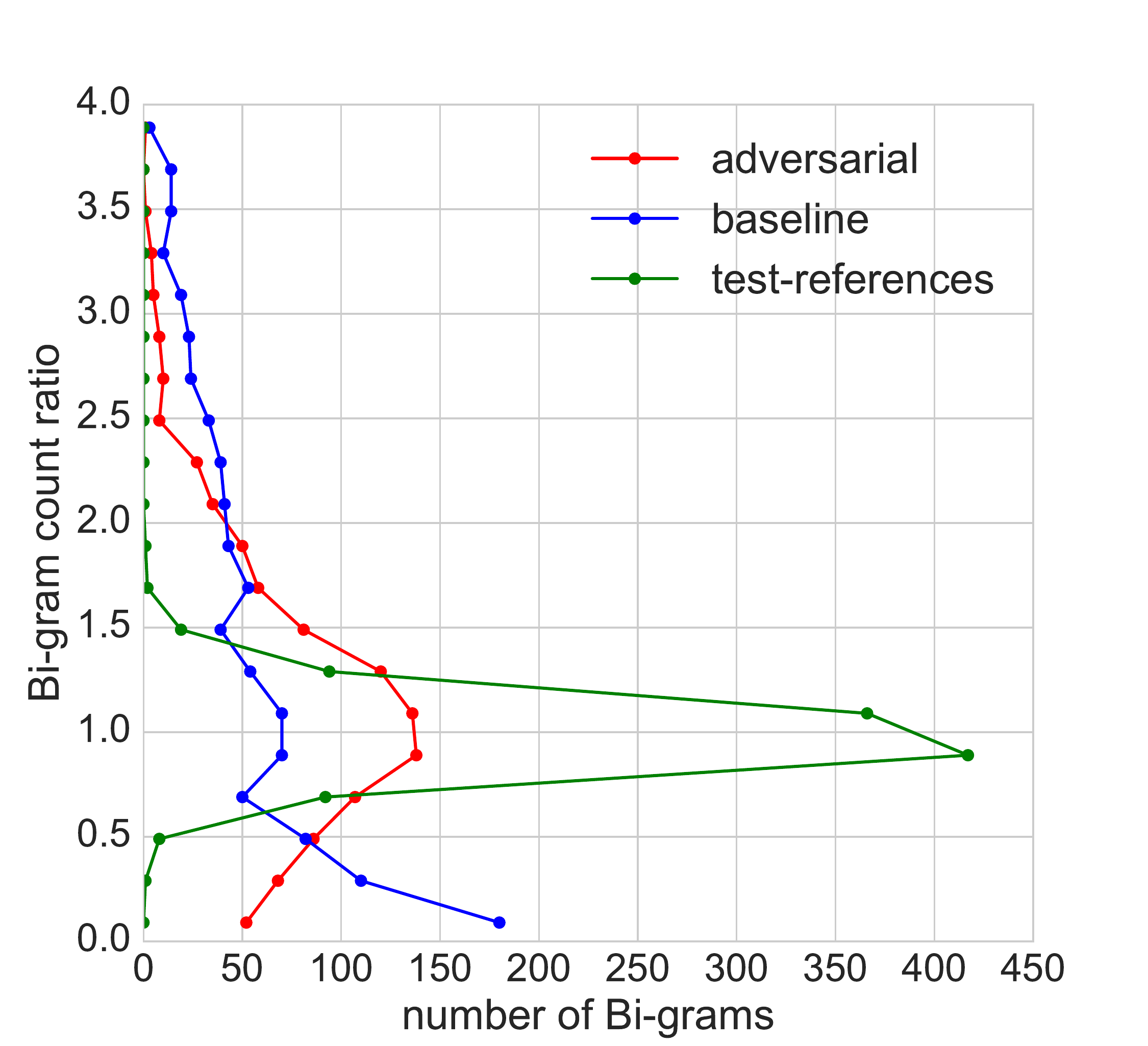}
    \includegraphics[width=0.49\columnwidth]{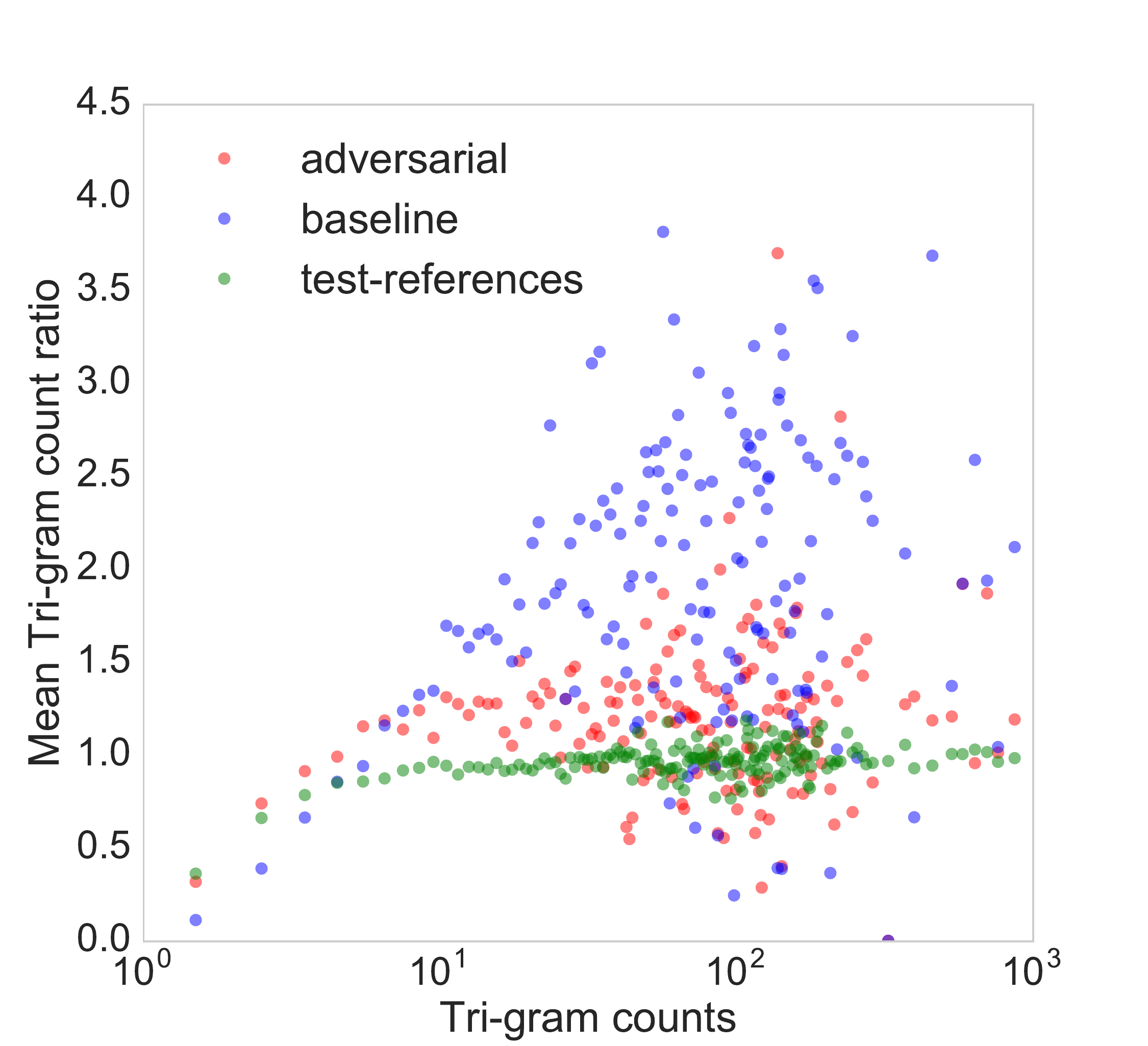}
    \includegraphics[width=0.49\columnwidth]{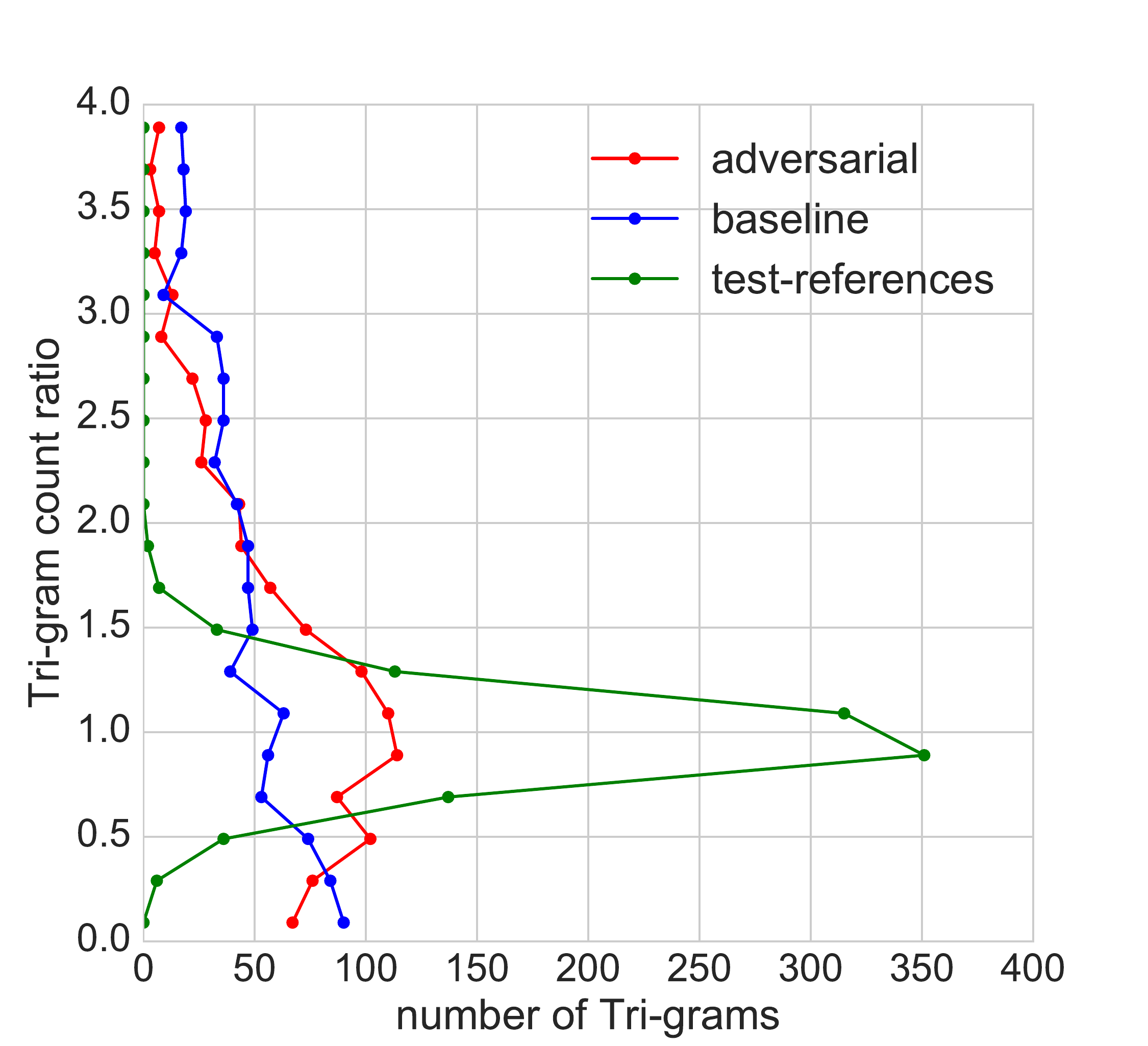}
    \caption{Comparison of $n$-gram count ratios in generated test-set captions by different models. Left side shows the mean $n$-gram count-ratios as a function of counts on training set. Right side shows the histogram of the count-ratios.
    }
    \label{fig:UnigramStats}
\end{figure}
To characterize how well the captions produced by the automatic methods match the statistics of the human written captions, we look at $n$-gram usage statistics in the generated captions. Specifically, we compute the ratio of the actual count of an
$n$-gram in the caption set produced by a model to the expected $n$-gram count based on the training data. 

Given that an $n$-gram occurred $m$ times in the
training set we can expect that it occurs $m*|\text{test-set}|/|\text{train-set}|$ times in the
test set. However actual counts may vary depending on how different the test set
is from the training set. We compute these ratios for reference captions in the test set to get an estimate of the expected variance of the count ratios. 
\begin{table}[tb]
\centering
\small
\setlength{\tabcolsep}{3pt}
\begin{tabular}{lcccccc}
\toprule
 &  &  &  &  & Vocab- & \% Novel  \\
 \bf Method & n &  Div-1 & Div-2 & mBleu-4 &ulary &  Sentences \\
\cmidrule(lr){1-1}\cmidrule(lr){2-2}\cmidrule(lr){3-5}\cmidrule(lr){6-7}
\multirow{2}{*}{Base-bs} & 1 of 5 & -- & -- & -- & 756  & 34.18 \\
							  & 5 of 5 & 0.28 & 0.38 & 0.78 &1085 & 44.27 \\
                              \cmidrule(lr){2-7}
\multirow{2}{*}{Base-samp} & 1 of 5 & -- & -- & -- & 839 & 52.04 \\
							   & 5 of 5 & 0.31 & 0.44 & 0.68 & 1460 & 55.24 \\
                               \cmidrule(lr){2-7}
\multirow{2}{*}{Adv-bs} & 1 of 5 & -- & -- & -- & 1508 & 68.62 \\
								& 5 of 5 & 0.34 & 0.44 & 0.70 & 2176 & 72.53  \\
                                \cmidrule(lr){2-7}
\multirow{2}{*}{Adv-samp} & 1 of 5 & -- & -- & -- & 1616 & 73.92 \\
								  & 5 of 5 & \bf 0.41 &\bf 0.55 &\bf 0.51 &\bf 2671 & \bf79.84 \\
\cmidrule(lr){1-7}
Human  & 1 of 5 & -- & -- & -- &  3347 & 92.80 \\
				captions				  & 5 of 5 & 0.53 & 0.74 & 0.20 & 7253 & 95.05 \\                                  
 \bottomrule
\end{tabular}
\caption{Diversity Statistics described in \Secref{sec:divstatsdef}. Higher values correspond to more diversity in all except mBleu-4, where lower is better.}
\label{tab:diversityStat}
\end{table}
\begin{figure}[htp]
\scriptsize
\begin{tabular}{p{0.2cm}p{2.4cm}p{2.25cm}p{2.3cm}}
  & \includegraphics[width=0.3\columnwidth, height=0.25\columnwidth]{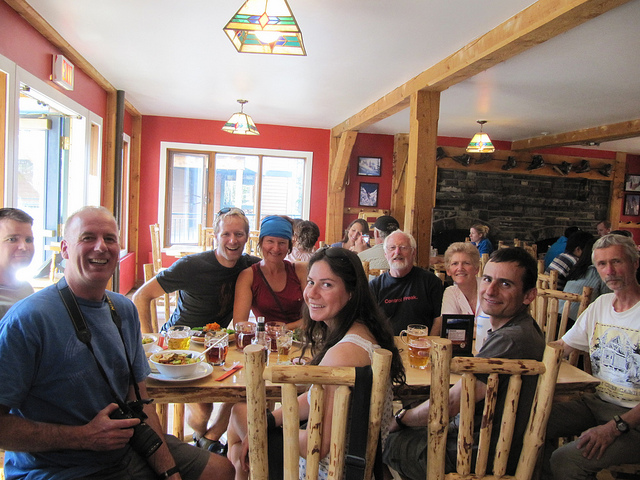}
  & \includegraphics[width=0.3\columnwidth, height=0.25\columnwidth]{images/COCO_val2014_000000244737.jpg}
  & \includegraphics[width=0.3\columnwidth, height=0.25\columnwidth]{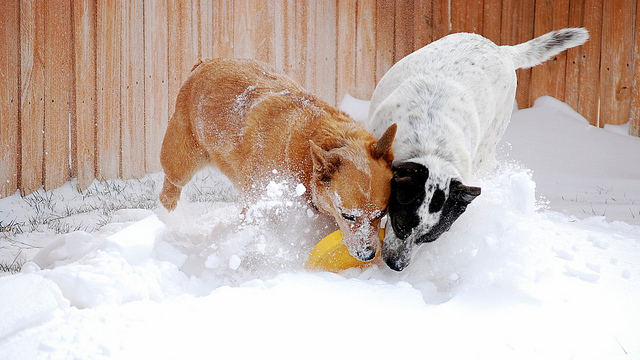}
  \\
  \textbf{Adv-bs} & a group of friends enjoying a dinner at the restauarant & several cows in their pen at the farm& A dog is trying to get something out of the snow\\
  \textbf{Base-bs}& a group of people sitting around a wooden table& a herd of cattle standing next to each other & a couple of dogs that are in the snow\\
\end{tabular}
\caption{Some qualitative examples comparing comparing captions generated by the our model to the baseline model.}
  \label{fig:qualsamps}
\end{figure}
The left side of Figure~\ref{fig:UnigramStats} shows the mean count ratios for uni-, bi- and tri-grams in the captions generated on test-set plotted against occurrence counts
in the training set. Histogram of these ratios are shown on the right side.

Count ratios for the reference captions from the test-set are shown in green. We see that the $n$-gram counts match well between the
training and test set human captions and the count ratios are spread
around $1.0$ with a small variance. 

The baseline model shows a clear bias towards more frequently occurring n-grams.
It consistently overuses more frequent n-grams~(ratio$>$1.0) from the training
set and under-uses less frequent ones~(ratio$<$1.0).  This trend is seen in all
the three plots, with more frequent tri-grams particularly prone to overuse.
It can also be observed in the histogram plots of the count ratios, that the
baseline model does a poor job of matching the statistics of the test set.

Our adversarial model does a much better job in matching these statistics. The histogram of the uni-gram count ratios are clearly closer to that of test reference captions. It does not seem to be significantly overusing the popular words, but there is still a trend of under utilizing some of the rarer words. It is however clearly better than the baseline model in this aspect. The improvement is less pronounced with the bi- and tri-grams, but still present.

Another clear benefit from using the adversarial training is observed in terms of diversity in the captions produced by the model. The diversity in terms of
both global statistics and per image diversity statistics is much higher in captions produced by the adversarial models compared to the baseline models.
This result is presented in \Tableref{tab:diversityStat}. We can see that the vocabulary size approximately doubles from 1085 in the baseline model to 2176  in the adversarial model using beamsearch. A similar trend is also seen comparing the
sampling variants.  As expected more diversity is achieved when sampling from the adversarial model instead of using beamsearch with vocabulary size increasing to 2671 in \emph{Adv-samp}. The effect of this increased diversity can be in the qualitative examples shown in \Figref{fig:qualsamps}. More qualitative samples are included in the supplementary material.

We can also see that the adversarial model learns to construct significantly more novel sentences compared to the baseline model with \emph{Adv-bs} producing novel captions 72.53\% of the time compared to  just 44.27\% by the \emph{beam-bs}.
All three per-image diversity statistics also
improve in the adversarial models indicating that they can produce a more diverse set of captions for any input image.

\Tableref{tab:diversityStat} also shows the diversity statistics on the reference captions on the test set. This
shows that although adversarial models do considerably better than
the baseline, there is still a gap in diversity statistics when compared to
the human written captions, especially in vocabulary size. 

\begin{figure}
\centering
    \includegraphics[width=0.6\columnwidth]{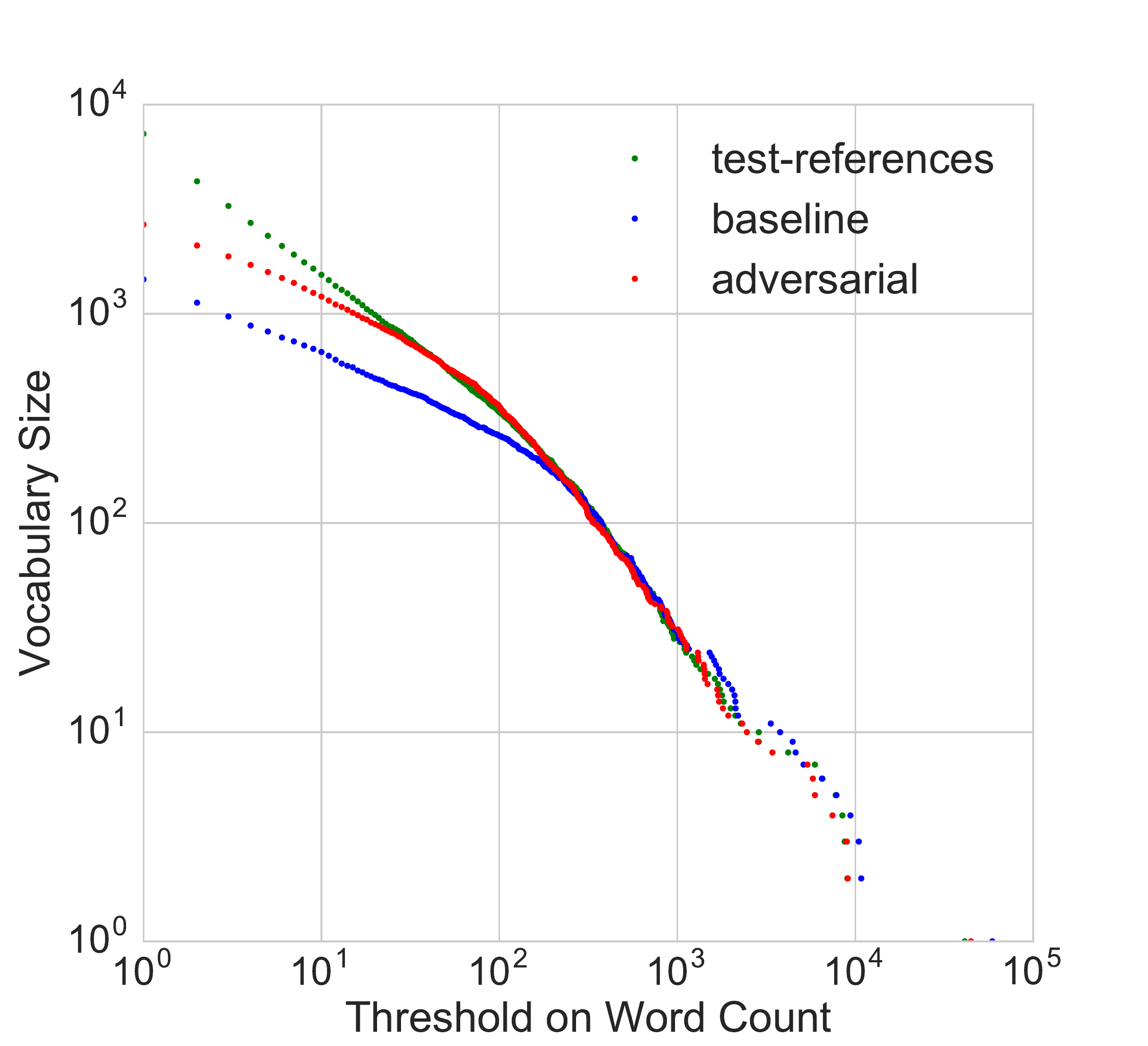}
   \caption{Vocabulary size as a function of word counts.}
\label{fig:Vocabularysize}
\end{figure}

Finally, \Figref{fig:Vocabularysize} plots the vocabulary size as a function of word count threshold, k. We see that the curve for the adversarial model better matches the human written captions compared to the baseline for all values of k. This illustrates that the gains in vocabulary size in adversarial models does not arise from using words with specific frequency, but is instead distributed evenly across word frequencies.

\subsection{Ablation Study}
\label{sec:result:ablation}
We conducted experiments to understand the importance of different components of
our architecture. The results are presented in
\Tableref{tab:intermed}. The baseline model for this experiment uses VGG~\cite{simonyan15iclr} features as $x_p$ input and is trained using maximum likelihood loss and is shown in the first row of \Tableref{tab:intermed}. The other four models use adversarial training. 

Comparing rows 1 and 2 of \Tableref{tab:intermed}, we see that adversarial training with a discriminator evaluating a single caption does badly. Both the diversity and Meteor score drop compared to the baseline. In this setting the generator can get away with producing one good caption~(mode collapse) for an image as the discriminator is unable to penalize the lack of
diversity in the generator. 

However, comparing rows 1 and 3, we see that adversarial training using a discriminator evaluating 5 captions simultaneously does much better in terms of Div-2 and vocabulary size. Adding feature matching loss further improves the diversity and also slightly improves accuracy in terms of Meteor score. Thus simultaneously evaluating multiple captions and using feature matching loss allows us to alleviate mode collapse generally observed in GANs.

Upgrading to the \emph{ResNet}\cite{he16cvpr} increases the Meteor score greatly and slightly increases the vocabulary size. \emph{ResNet} features provide richer visual information which is used by the generator to produce diverse but still correct captions.

We also notice that the generator learns to ignore the input noise. This is because there is sufficient stochasticity in the generation process due to sequential sampling of words and thus the generator doesn't need the additional noise input to increase output diversity. Similar observation was reported in other conditional GAN architectures~\cite{isola2016image, mathieu2015deep}

\begin{table}[tb]
\centering
\small
\begin{tabular}{p{1cm}p{1cm}p{1.3cm}p{1cm}p{0.8cm}p{0.8cm}p{1cm}}
\toprule
Image Feature&{Evalset size (p)}
&Feature Matching& Meteor &
Div-2&Vocab. Size\\
 \cmidrule(lr){1-3}\cmidrule(lr){4-6}
 \multicolumn{3}{l}{VGG baseline}& 0.247 & 0.44 & 1367\\ 
 \cmidrule(lr){1-3}\cmidrule(lr){4-6}%
 VGG & 1 & No & 0.179  & 0.40 & 812\\
 VGG & 5 & No & 0.197 & 0.52 & 1810 \\
 VGG & 5 & yes & 0.207 &\bf0.59 & 2547\\
 ResNet & 5 & yes &\bf0.236 & 0.55 &\bf 2671\\
 \bottomrule
\end{tabular}
\caption{Performance comparison of various configurations of the adversarial caption generator on the validation set.}
\label{tab:intermed}
\vspace{-1em}
\end{table}

\section{Conclusions}
We have presented an adversarial caption generator model which is explicitly trained to generate diverse captions for images. 
We achieve this by utilizing a discriminator network designed to promote diversity and use the adversarial learning framework to train our generator. Results show that our adversarial model produces captions which are diverse and match the statistics of human generated captions significantly better than the baseline model. The adversarial model also uses larger vocabulary and is able to produce significantly more novel captions. The increased diversity is achieved while preserving accuracy of the generated captions, as shown through a human evaluation.

\section*{Acknowledgements}
This research was supported by the German Research Foundation (DFG CRC 1223) and by the Berkeley Artificial Intelligence Research (BAIR) Lab.

\small
\bibliographystyle{ieee}

\bibliography{biblioLong,related,rohrbach,paper}

\clearpage
\section{Supplementary Material}
We present several qualitative examples to illustrate the strengths of our adversarially trained caption generator. 
All the examples are from the sampled versions of the adversarial~(\emph{adv-samp}) and the baseline~(\emph{base-samp}) models presented above. %
We show qualitative examples to highlight two main merits of the adversarial caption generator. First, we demonstrate diversity when sampling multiple captions for each image in \secref{sec:appx1}. Next, we illustrate diversity across images in \secref{sec:appx2}.

\subsection{Illustrating diversity in captions for each image}
\label{sec:appx1}
\begin{figure*}[htp]
\begin{center}
\centering
\footnotesize
\begin{tabular}{x{3mm}x{5.4cm}x{5.7cm}x{5.4cm}}
  &\includegraphics[width=0.55\columnwidth, height=0.45\columnwidth]{images/COCO_val2014_000000556349.jpg}
  &\includegraphics[width=0.55\columnwidth, height=0.45\columnwidth]{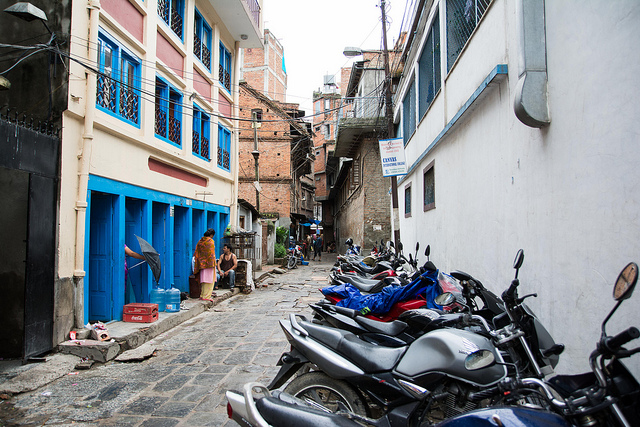}
  &\includegraphics[width=0.55\columnwidth, height=0.45\columnwidth]{images/COCO_val2014_000000230008.jpg}
  \\
  \cmidrule(lr){2-2}  \cmidrule(lr){3-3} \cmidrule(lr){4-4}
  \multirow{3}{*}{\parbox[c][][c]{3mm}{Adv-samp}} & \mybullet \red{a red} motorcycle parked \red{on the side of the} road& a motor cycle parked outside \red{a building} with people nearby& a motorcycle parked \red{in front of a group of people}\\
                            & \mybullet a motorcycle is parked \red{on a city} street& \mybullet a motor bike parked \red{in front of a building}& a police officer \red{on a} motorcycle \red{in front of a} crowd \red{of people}\\
                            & \red{a red} motorcycle parked \red{on a street in a city}& a row of bicycles parked outside \red{of a building}& a police officer on his motorcycle \red{in front of a} crowd\\
  \cmidrule(lr){2-2}  \cmidrule(lr){3-3} \cmidrule(lr){4-4}
  \multirow{3}{*}{\parbox[c][][c]{3mm}{Base-samp}} & \mybullet \red{a man} \red{riding a} motorcycle \red{down a street}&\mybullet \red{a group of people} riding bikes \red{down a street} &\mybullet a motorcycle parked \red{on the side of a} road\\
                             & \mybullet \red{a person} \red{riding a} motorcycle \red{on a city} street& \red{a group of people on a street} with motorcycles &a motorcycle is parked \red{on the side of a} road\\
                             & \mybullet a motorcycle is parked \red{on the side of the} road& \red{a group of people on a street} with motorcycles &a motorcycle parked \red{on the side of a} road \red{with a person} walking by\\
\\
  &\includegraphics[width=0.55\columnwidth, height=0.45\columnwidth]{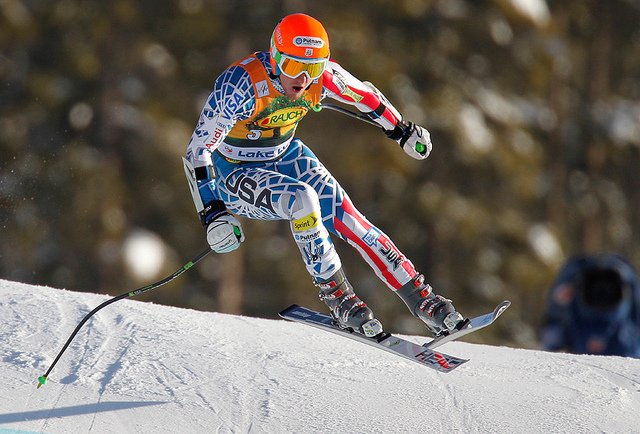}
  &\includegraphics[width=0.55\columnwidth, height=0.45\columnwidth]{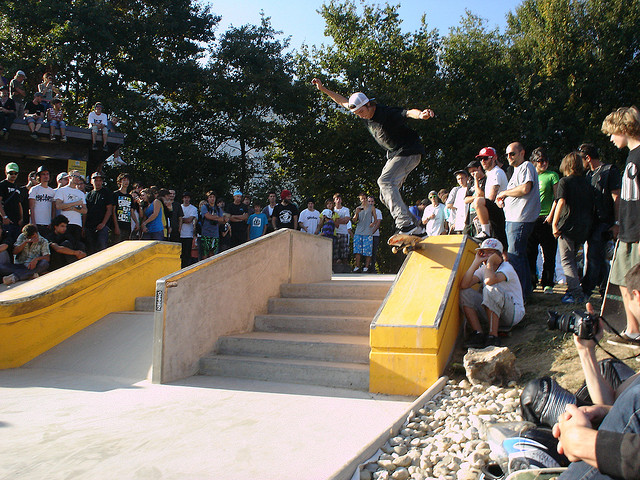}
  &\includegraphics[width=0.55\columnwidth, height=0.45\columnwidth]{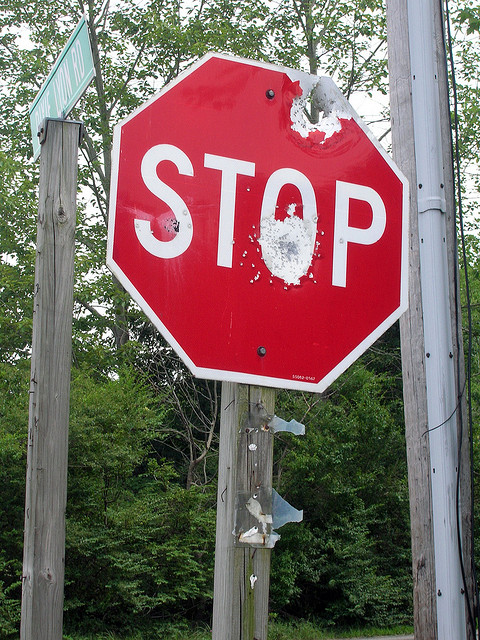}
  \\
  \cmidrule(lr){2-2}  \cmidrule(lr){3-3} \cmidrule(lr){4-4}
  \multirow{3}{*}{\parbox[c][][c]{3mm}{Adv-samp}} & a skier is jumping over a snow covered hill& \red{a group of people} watching a skateboarder do stunts & \mybullet a stop sign with graffiti written \red{on it}\\
                            & \red{a person} on skis jumping over a hill& \red{a group of} skateboarders performing tricks \red{at a} skate park& a stop sign \red{with a} few stickers \red{on it}\\
                            & a skier is in mid air after completing a jump& \red{a group of} skateboarders watch as others watch & a stop sign has graffiti written all over it\\
  \cmidrule(lr){2-2}  \cmidrule(lr){3-3} \cmidrule(lr){4-4}
  \multirow{3}{*}{\parbox[c][][c]{3mm}{Base-samp}}& \mybullet \red{a man} riding skis \red{down a} snow covered slope&\mybullet \red{a man} doing a trick \red{on a skateboard at a} skate park&a stop sign \red{with a street} sign \red{on top of} it\\
                             & \mybullet \red{a man} riding skis \red{down a} snow covered slope &\mybullet \red{a man} riding \red{a skateboard down a} rail &a stop sign \red{with a street} sign \red{on top of} it\\
                             & \red{a person on a} snowboard jumping over a ramp &\mybullet \red{a man} doing a trick \red{on a skateboard in a} park &\mybullet a stop sign \red{with a street} sign above it\\
\\
  &\includegraphics[width=0.55\columnwidth, height=0.45\columnwidth]{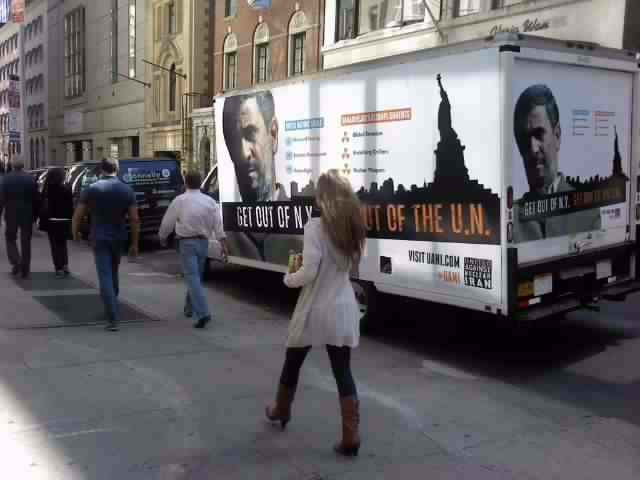}
  &\includegraphics[width=0.55\columnwidth, height=0.45\columnwidth]{images/COCO_val2014_000000213224.jpg}
  &\includegraphics[width=0.55\columnwidth, height=0.45\columnwidth]{images/COCO_val2014_000000069795.jpg}
  \\
  \cmidrule(lr){2-2}  \cmidrule(lr){3-3} \cmidrule(lr){4-4}
  \multirow{3}{*}{\parbox[c][][c]{3mm}{Adv-samp}} & \red{a man standing next to a} mail truck & a bouquet of flowers \red{in a} vase \red{on a table}& a plant \red{in a} vase \red{on a wooden} porch\\
                            & \red{a picture of people} standing outside a business& a bouquet of flowers \red{in a} vase \red{on a table}& \red{a small} blue flower vase \red{sitting on a wooden} porch\\
                            & \red{a couple of people} standing \red{by a} bus & a vase full of purple flowers \red{sitting on a table}& \red{a large} flower arrangement \red{in a} vase \red{on a} corner\\
  \cmidrule(lr){2-2}  \cmidrule(lr){3-3} \cmidrule(lr){4-4}
  \multirow{3}{*}{\parbox[c][][c]{3mm}{Base-samp}} & \red{a group of people} standing around \red{a white} truck& \mybullet a vase \red{filled with} flowers \red{sitting on top of a table}&\mybullet a vase with flowers in it \red{sitting on a table}\\
                             &\red{a group of people} standing around \red{a white} truck & \mybullet a vase \red{filled with} flowers \red{sitting on top of a table} &\mybullet a vase of flowers \red{sitting on a table}\\
                             & \red{a group of people standing on a street next to a white} truck& \mybullet a vase of flowers \red{sitting on a table}&\mybullet a vase with flowers in it \red{on a table}\\
\end{tabular}
\vspace{-0.4cm}
\caption{Comparing 3 captions sampled from adversarial model to the baseline model. Bi-grams which are top-100 frequent bi-grams in the training set are highlighted in red. Captions which are replicas from training set are marked with a \mybullet.
}
  \label{fig:appx3}
\end{center}
\end{figure*}

\begin{figure*}[thp]
\begin{center}
\footnotesize
\begin{tabular}{x{3.0mm}x{5.4cm}x{5.7cm}x{5.3cm}}
  &\includegraphics[width=0.55\columnwidth, height=0.45\columnwidth]{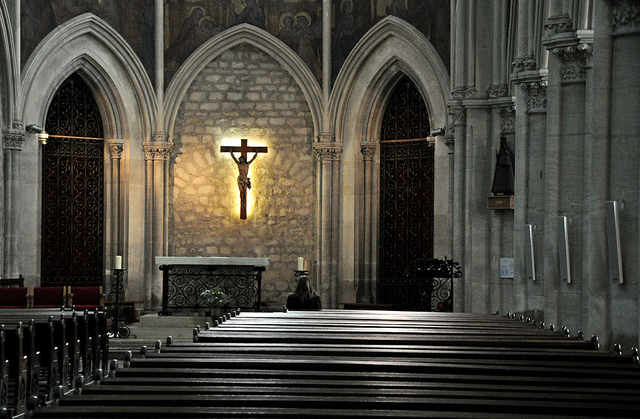}
  &\includegraphics[width=0.55\columnwidth, height=0.45\columnwidth]{images/COCO_val2014_000000094157.jpg}
  &\includegraphics[width=0.55\columnwidth, height=0.45\columnwidth]{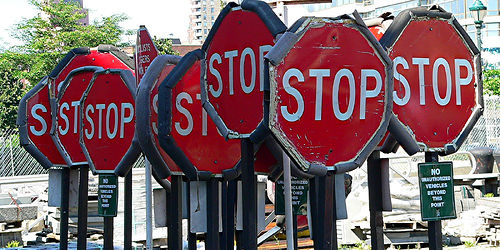}
  \\
  \cmidrule(lr){2-2}  \cmidrule(lr){3-3} \cmidrule(lr){4-4}
  \multirow{3}{*}{\parbox[c][][c]{3mm}{Adv-samp}} & a long line of stairs leading \red{to a} church& \red{a large} church \red{with a} very tall tower & several stop signs \red{in front of} some buildings\\
                            & \mybullet \red{a large} cathedral \red{filled with} lots of pews& \red{a large} tall brick building \red{with a} clock on it& a stop sign \red{in front of} some graffiti writing\\
                            & a cathedral with stained glass windows \red{and a} few people& a church steeple \red{with a} clock on its side& several stop signs lined up \red{in a} row\\
  \cmidrule(lr){2-2}  \cmidrule(lr){3-3} \cmidrule(lr){4-4}
  \multirow{3}{*}{\parbox[c][][c]{3mm}{Base-samp}} & \red{a large} building \red{with a large} window \red{and a building}& \red{a large} clock tower \red{in a city}& \mybullet a stop sign with graffiti \red{on it} \\
                             & a row of benches \red{in front of a building}& \red{a large} clock tower \red{in a city with a} sky background& a stop sign is shown \red{with a} lot of graffiti \red{on it}\\
                             & a church \red{with a large} window \red{and a large} building& \mybullet \red{a large} clock tower \red{in the} middle \red{of a} park& a stop sign \red{and a} stop sign \red{in front of a building}\\
\\
  &\includegraphics[width=0.55\columnwidth, height=0.45\columnwidth]{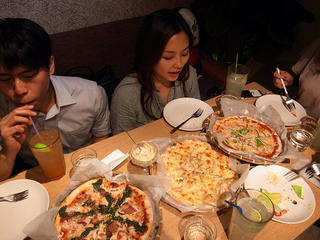}
  &\includegraphics[width=0.55\columnwidth, height=0.45\columnwidth]{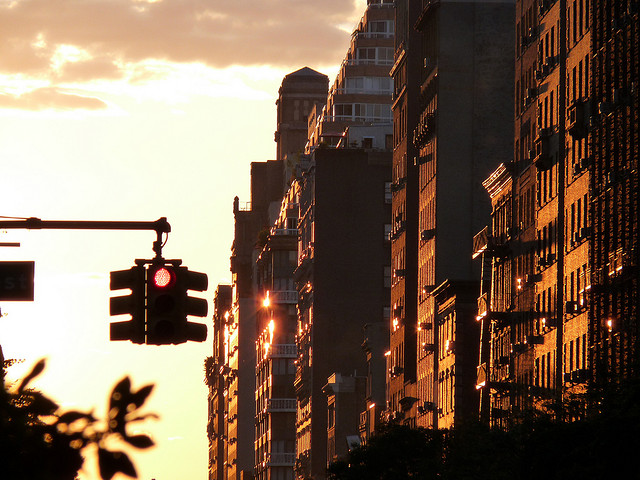}
  &\includegraphics[width=0.55\columnwidth, height=0.45\columnwidth]{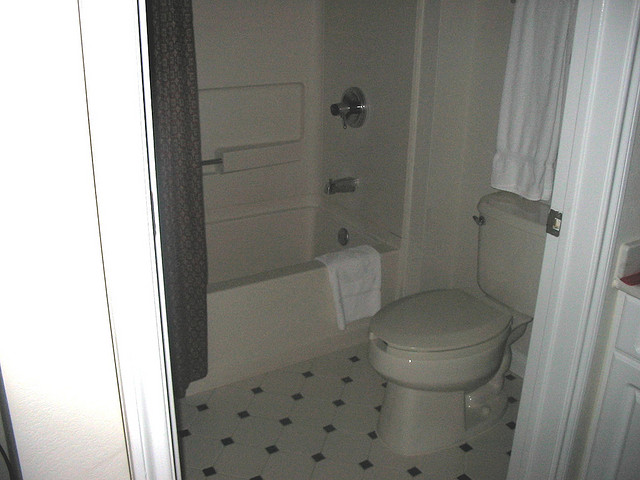}\\
  \cmidrule(lr){2-2}  \cmidrule(lr){3-3} \cmidrule(lr){4-4}
  \multirow{3}{*}{\parbox[c][][c]{3mm}{Adv-samp}}  & a family enjoying pizza \red{at a} restaurant party& a view \red{of a} city street at dusk& \red{a white} toilet sitting underneath a shower curtain  \\
                                                   & \red{a group of} friends enjoying pizza and drinking beer & a city street with many buildings and buildings & \mybullet \red{a white} toilet sitting underneath \red{a bathroom} window\\
                                                   & \red{a group of} kids enjoying pizza and drinking beer & a view \red{of a} city intersection \red{in the} evening & a bathroom \red{with a} shower curtain open \red{and a} toilet in it\\
  \cmidrule(lr){2-2}  \cmidrule(lr){3-3} \cmidrule(lr){4-4}
  \multirow{3}{*}{\parbox[c][][c]{3mm}{Base-samp}}  &\red{a group of people} sitting \red{at a table with} pizza&a traffic light \red{in a city} with tall buildings& \mybullet \red{a bathroom with a} toilet \red{and a} shower\\
                                                    &\red{a group of people} sitting \red{at a table with} pizza and drinks  &a traffic light \red{in a city} with tall buildings& \mybullet \red{a bathroom with a} toilet \red{and a} shower \\
                                                    &\red{a group of people} sitting \red{at a table with} pizza &\mybullet a traffic light \red{and a street} sign \red{on a city} street&\red{a white} toilet sitting \red{next to a} shower \red{in a bathroom}\\
\end{tabular}
\caption{Comparing 3 captions sampled from adversarial model to the baseline model. %
}
  \label{fig:appx2}
\end{center}
\end{figure*}

To qualitatively demonstrate the diverse captions produced by the our adversarial model for each image, we visualize three captions produced by the adversarial and the baseline model for each input image.
This is shown in figures~\ref{fig:appx3} and \ref{fig:appx2}.
The captions are obtained by retaining the top three caption samples out of five (ranked by models' probability) from each model. 
Here bi-grams which are top-100 frequent bi-grams in the training set are highlighted in red (e.g., ``a group'' and ``group of''). 
Additionally captions which are replicas from training set are marked with a `\mybullet' symbol.
We can see that adversarial generator produces more diverse sets of captions for each image without over-using more frequent bi-grams and producing more novel sentences.
For example, in the two figures, we see that the baseline model produces 22 captions (out of 45) which are copies from the training set, whereas the adversarial model does so only six times.


\subsection{Illustrating diversity across images}
\label{sec:appx2}
\begin{figure*}[htp]
\begin{center}
\centering
\small
\begin{tabular}{x{5mm}x{3cm}x{3cm}x{3cm}x{3cm}x{3cm}}
  &\includegraphics[width=0.40\columnwidth, height=0.35\columnwidth]{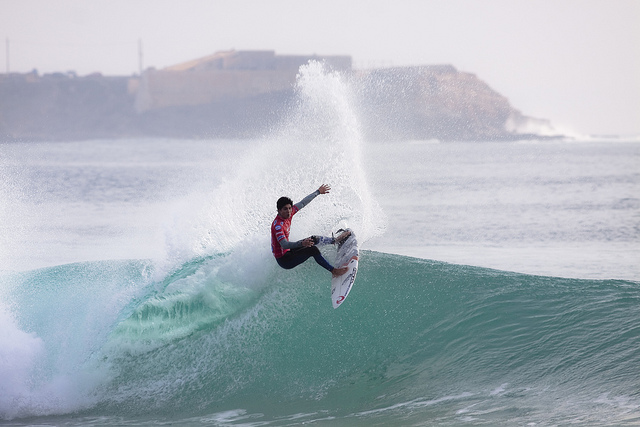}
  &\includegraphics[width=0.40\columnwidth, height=0.35\columnwidth]{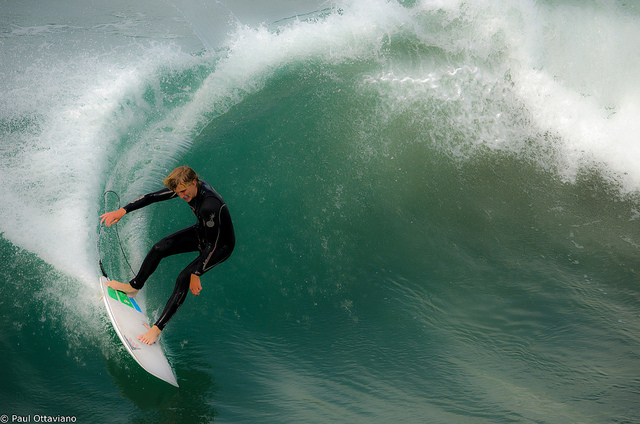}
  &\includegraphics[width=0.40\columnwidth, height=0.35\columnwidth]{images/COCO_val2014_000000381360.jpg}
  &\includegraphics[width=0.40\columnwidth, height=0.35\columnwidth]{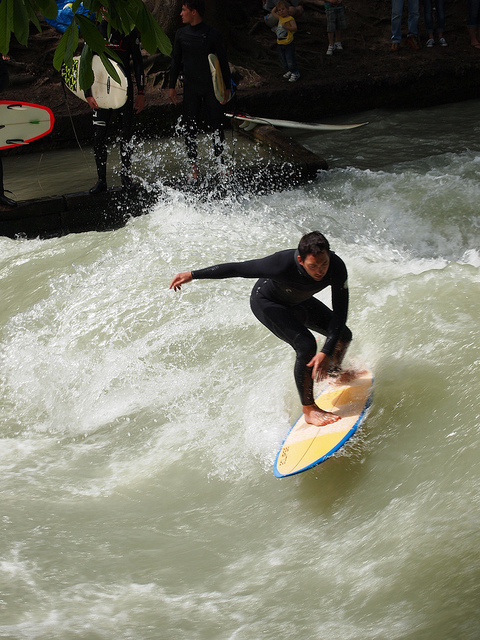}
  &\includegraphics[width=0.40\columnwidth, height=0.35\columnwidth]{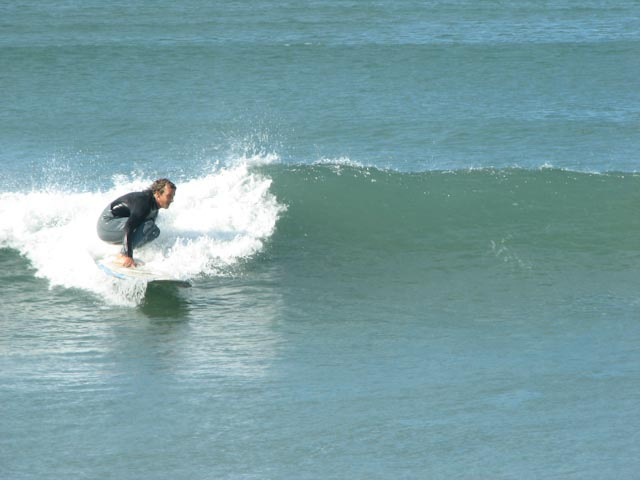}
  \\
  \multirow{1}{*}{\parbox[c][][c]{5mm}{Adv-samp}} & a surfer rides a large wave in the ocean& a surfer is falling off his board as he rides a wave & a person on a surfboard riding a wave& a man surfing on a surfboard in rough waters & a surfer rides a small wave in the ocean\\
  \\
  \multirow{1}{*}{\parbox[c][][c]{5mm}{Base-samp}} & \multicolumn{5}{c}{$\cdots\cdots\cdots\cdots\cdots$\quad\quad a man riding a wave on top of a surfboard \quad\quad$\cdots\cdots\cdots\cdots\cdots$ }\\

  &\includegraphics[width=0.40\columnwidth, height=0.35\columnwidth]{images/COCO_val2014_000000300034.jpg}
  &\includegraphics[width=0.40\columnwidth, height=0.35\columnwidth]{images/COCO_val2014_000000255401.jpg}
  &\includegraphics[width=0.40\columnwidth, height=0.35\columnwidth]{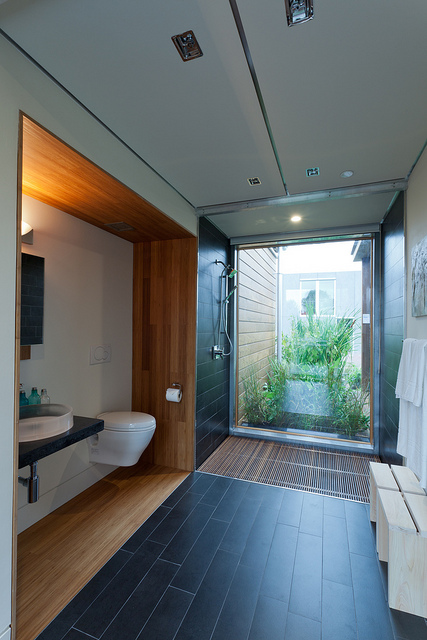}
  &\includegraphics[width=0.40\columnwidth, height=0.35\columnwidth]{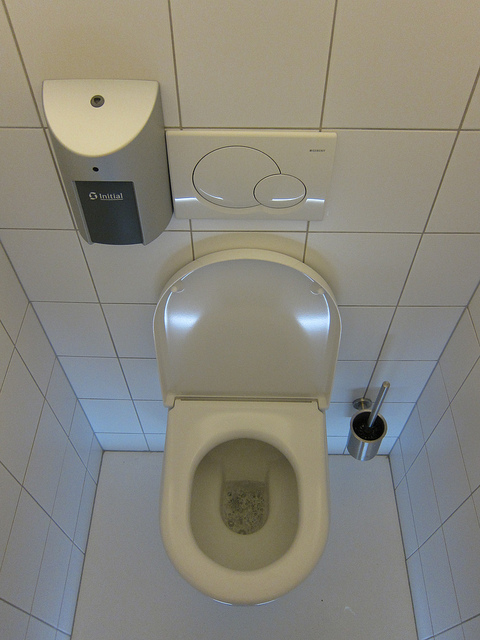}
  &\includegraphics[width=0.40\columnwidth, height=0.35\columnwidth]{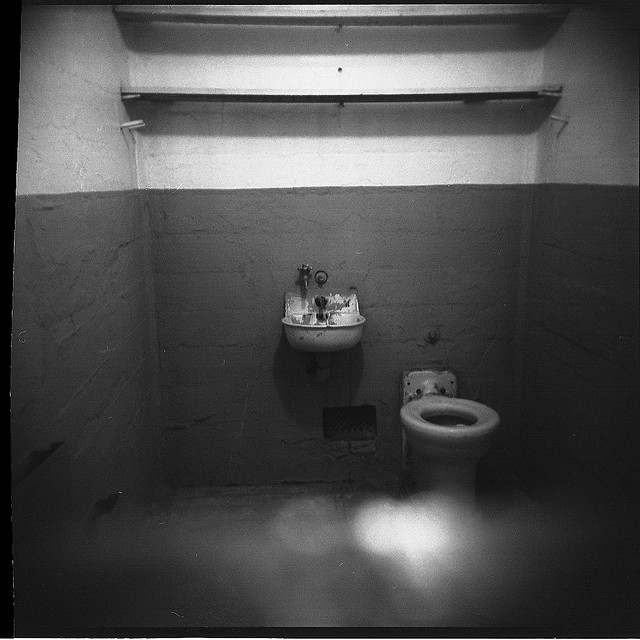}
  \\
  \multirow{1}{*}{\parbox[c][][c]{5mm}{Adv-samp}} & a bathroom with a walk in shower and a sink & a dirty bathroom with a broken toilet and sink& a view of a very nice looking rest room& a white toilet in a public restroom stall& a small bathroom has a broken toilet and a broken sink\\
  \\
  \multirow{1}{*}{\parbox[c][][c]{5mm}{Base-samp}} & \multicolumn{5}{c}{$\cdots\cdots\cdots\cdots\cdots$\quad\quad a bathroom with a toilet and a sink \quad\quad$\cdots\cdots\cdots\cdots\cdots$ }\\
  &\includegraphics[width=0.40\columnwidth, height=0.35\columnwidth]{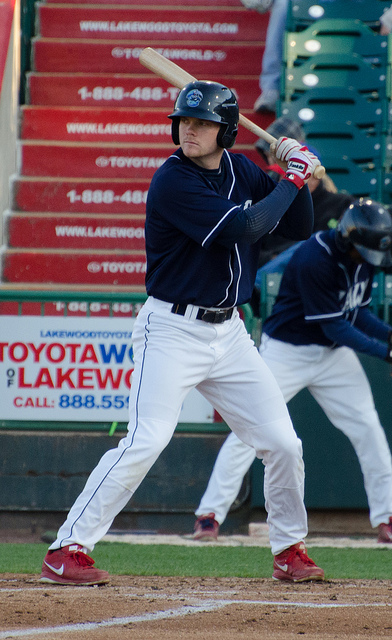}
  &\includegraphics[width=0.40\columnwidth, height=0.35\columnwidth]{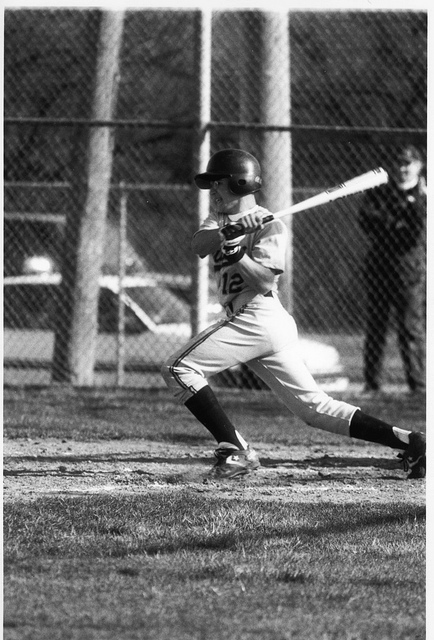}
  &\includegraphics[width=0.40\columnwidth, height=0.35\columnwidth]{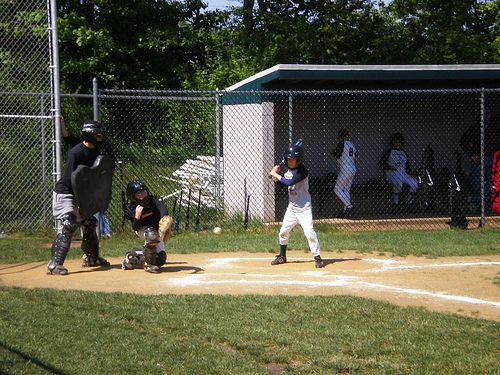}
  &\includegraphics[width=0.40\columnwidth, height=0.35\columnwidth]{images/COCO_val2014_000000320524.jpg}
  &\includegraphics[width=0.40\columnwidth, height=0.35\columnwidth]{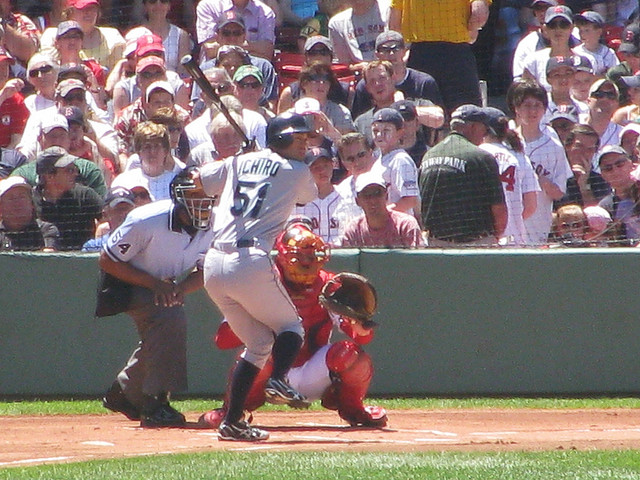}
  \\
  \multirow{1}{*}{\parbox[c][][c]{5mm}{Adv-samp}} & a baseball player getting ready to swing at a pitch & a boy in a baseball uniform swinging a bat& a group of kids playing baseball on a field& a baseball game in progress with the batter upt to swing & a crowd watches a baseball game being played\\
  \\
  \multirow{1}{*}{\parbox[c][][c]{5mm}{Base-samp}} & \multicolumn{5}{c}{$\cdots\cdots\cdots\cdots\cdots$\quad\quad a baseball player swinging a bat at a ball \quad\quad$\cdots\cdots\cdots\cdots\cdots$ }\\

  &\includegraphics[width=0.40\columnwidth, height=0.35\columnwidth]{images/COCO_val2014_000000225299.jpg}
  &\includegraphics[width=0.40\columnwidth, height=0.35\columnwidth]{images/COCO_val2014_000000481187.jpg}
  &\includegraphics[width=0.40\columnwidth, height=0.35\columnwidth]{images/COCO_val2014_000000038210.jpg}
  &\includegraphics[width=0.40\columnwidth, height=0.35\columnwidth]{images/COCO_val2014_000000539079.jpg}
  &\includegraphics[width=0.40\columnwidth, height=0.35\columnwidth]{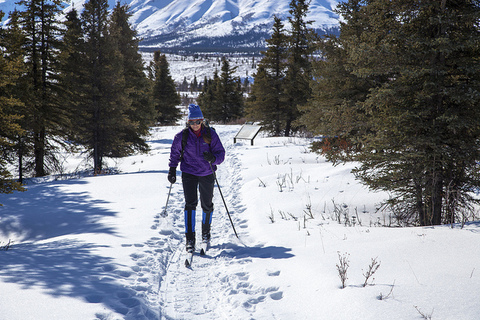}
  \\
  \multirow{1}{*}{\parbox[c][][c]{5mm}{Adv-samp}} & a person on skis jumping over a ramp& a skier is making a turn on a course& a cross country skier makes his way through the snow & a skier is headed down a steep slope & a person cross country skiing on a trail\\
  \\
  \multirow{1}{*}{\parbox[c][][c]{5mm}{Base-samp}} & \multicolumn{5}{c}{$\cdots\cdots\cdots\cdots\cdots$\quad\quad a man riding skis down a snow covered slope \quad\quad$\cdots\cdots\cdots\cdots\cdots$ }\\
\\
\end{tabular}
\caption{Illustrating diversity across images}
\label{fig:appx5}
\end{center}
\end{figure*}

\begin{figure*}[htp]
\begin{center}
\centering
\small
\begin{tabular}{x{5mm}x{3cm}x{3cm}x{3cm}x{3cm}x{3cm}}
  &\includegraphics[width=0.40\columnwidth, height=0.35\columnwidth]{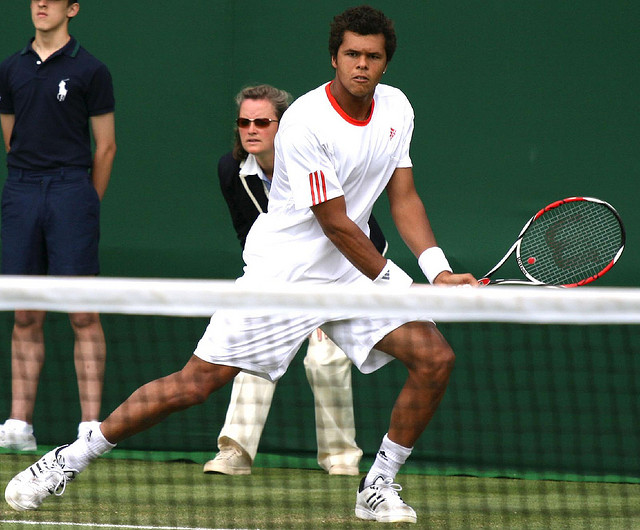}
  &\includegraphics[width=0.40\columnwidth, height=0.35\columnwidth]{images/COCO_val2014_000000294853.jpg}
  &\includegraphics[width=0.40\columnwidth, height=0.35\columnwidth]{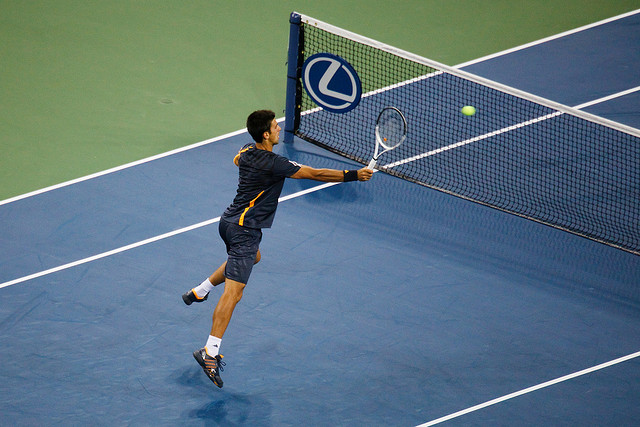}
  &\includegraphics[width=0.40\columnwidth, height=0.35\columnwidth]{images/COCO_val2014_000000379332.jpg}
  &\includegraphics[width=0.40\columnwidth, height=0.35\columnwidth]{images/COCO_val2014_000000376900.jpg}
  \\
  \multirow{1}{*}{\parbox[c][][c]{5mm}{Adv-samp}} & a tennis player gets ready to return a serve& two men dressed in costumes and holding tennis rackets& a tennis player hits the ball during a match& a male tennis player in action on the court& a man in white is about to serve a tennis ball\\
  \\
  \multirow{1}{*}{\parbox[c][][c]{5mm}{Base-samp}} & \multicolumn{5}{c}{$\cdots\cdots\cdots\cdots\cdots$\quad\quad a man holding a tennis racquet on a tennis court \quad\quad$\cdots\cdots\cdots\cdots\cdots$ }\\

  &\includegraphics[width=0.40\columnwidth, height=0.35\columnwidth]{images/COCO_val2014_000000519744.jpg}
  &\includegraphics[width=0.40\columnwidth, height=0.35\columnwidth]{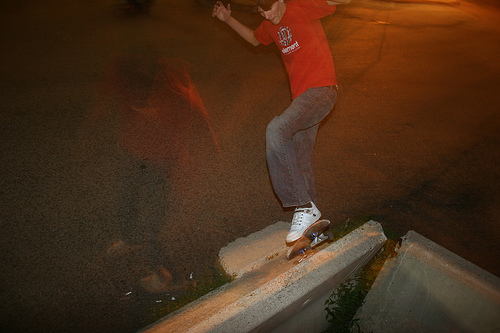}
  &\includegraphics[width=0.40\columnwidth, height=0.35\columnwidth]{images/COCO_val2014_000000229553.jpg}
  &\includegraphics[width=0.40\columnwidth, height=0.35\columnwidth]{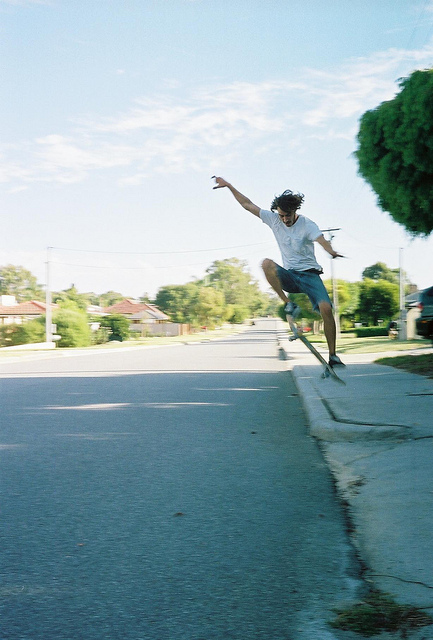}
  &\includegraphics[width=0.40\columnwidth, height=0.35\columnwidth]{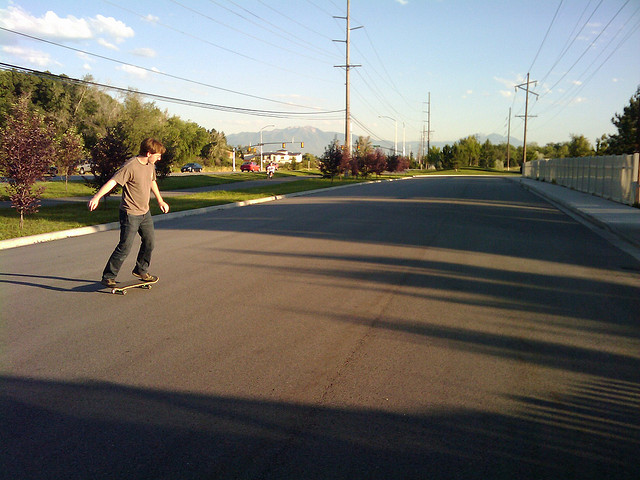}
  \\
  \multirow{1}{*}{\parbox[c][][c]{5mm}{Adv-samp}} & a young boy riding a skateboard down a street & a skateboarder is attempting to do a trick& a boy wearing a helmet and knee pads riding a skateboard& a boy in white shirt doing a trick on a skateboard& a boy is skateboarding down a street in a neighborhood \\
  \\
  \multirow{1}{*}{\parbox[c][][c]{5mm}{Base-samp}} & \multicolumn{5}{c}{$\cdots\cdots\cdots\cdots\cdots$\quad\quad a man riding a skateboard down a steet \quad\quad$\cdots\cdots\cdots\cdots\cdots$ }\\
  &\includegraphics[width=0.40\columnwidth, height=0.35\columnwidth]{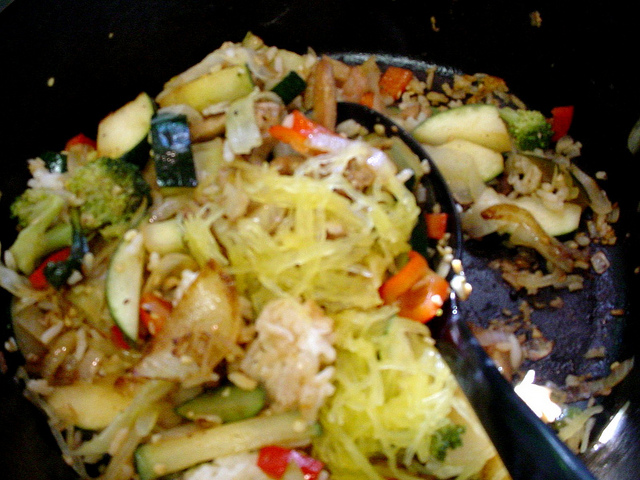}
  &\includegraphics[width=0.40\columnwidth, height=0.35\columnwidth]{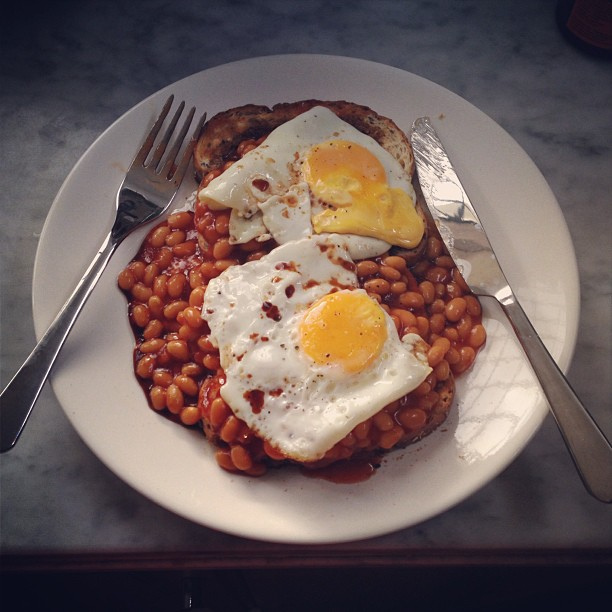}
  &\includegraphics[width=0.40\columnwidth, height=0.35\columnwidth]{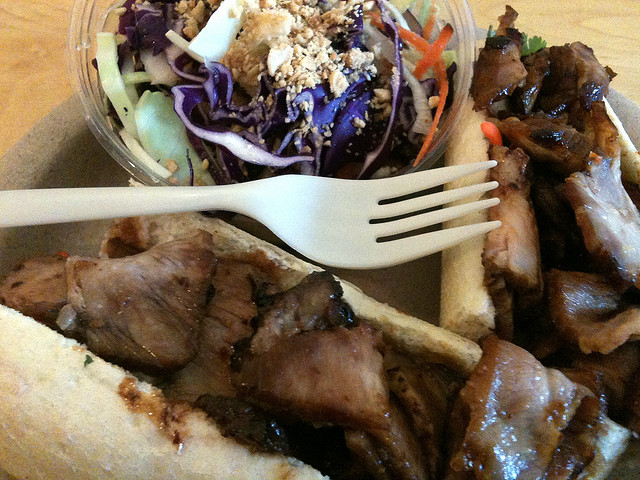}
  &\includegraphics[width=0.40\columnwidth, height=0.35\columnwidth]{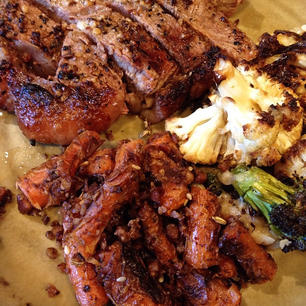}
  &\includegraphics[width=0.40\columnwidth, height=0.35\columnwidth]{images/COCO_val2014_000000249969.jpg}
  \\
  \multirow{1}{*}{\parbox[c][][c]{5mm}{Adv-samp}} &a dish with noodles and vegetables in it & a plate of food that has some fried eggs on it & a meal consisting of rice meat and vegetables & a close up of some meat and potatoes & a plate with some meat and vegetables on it\\
  \\
  \multirow{1}{*}{\parbox[c][][c]{5mm}{Base-samp}} & \multicolumn{5}{c}{$\cdots\cdots\cdots\cdots\cdots$\quad\quad a plate of food with meat and vegetables \quad\quad$\cdots\cdots\cdots\cdots\cdots$ }\\
  &\includegraphics[width=0.40\columnwidth, height=0.35\columnwidth]{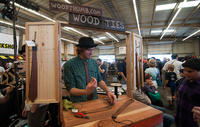}
  &\includegraphics[width=0.40\columnwidth, height=0.35\columnwidth]{images/COCO_val2014_000000363188.jpg}
  &\includegraphics[width=0.40\columnwidth, height=0.35\columnwidth]{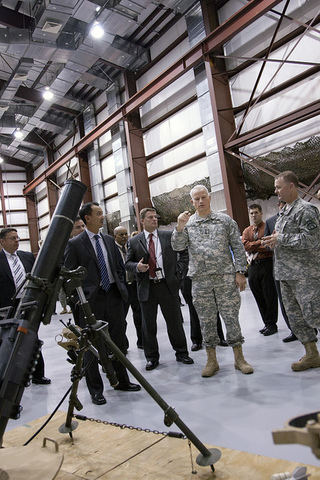}
  &\includegraphics[width=0.40\columnwidth, height=0.35\columnwidth]{images/COCO_val2014_000000357978.jpg}
  &\includegraphics[width=0.40\columnwidth, height=0.35\columnwidth]{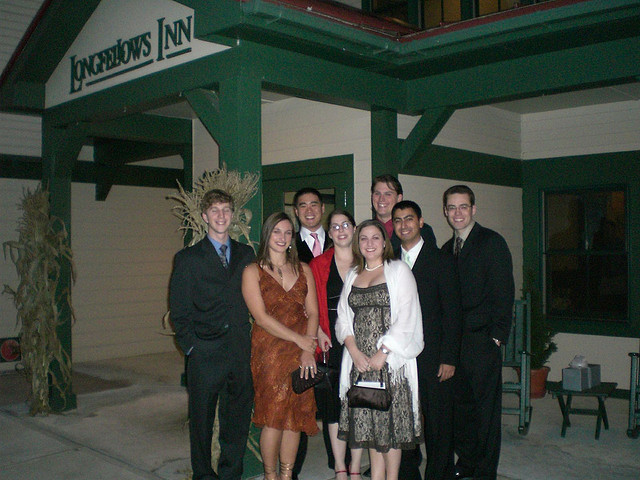}
  \\
  \multirow{1}{*}{\parbox[c][][c]{5mm}{Adv-samp}} & a group of people standing around a shop & a group of young people standing around talking on cell phones& a group of soldiers stand in front of microphones& a couple of women standing next to a man in front of a store& a group of people posing for a photo in formal wear\\
  \\
  \multirow{1}{*}{\parbox[c][][c]{5mm}{Base-samp}} & \multicolumn{5}{c}{$\cdots\cdots\cdots\cdots\cdots$\quad\quad a group of people standing around a table \quad\quad$\cdots\cdots\cdots\cdots\cdots$ }\\
\end{tabular}
\caption{Illustrating diversity across images}
\label{fig:appx6}
\end{center}
\end{figure*}

\begin{figure*}[bh]
\begin{center}
\centering
\footnotesize
\begin{tabular}{x{5mm}x{3cm}x{3cm}x{3cm}x{3cm}x{3cm}}
  &\includegraphics[width=0.40\columnwidth, height=0.35\columnwidth]{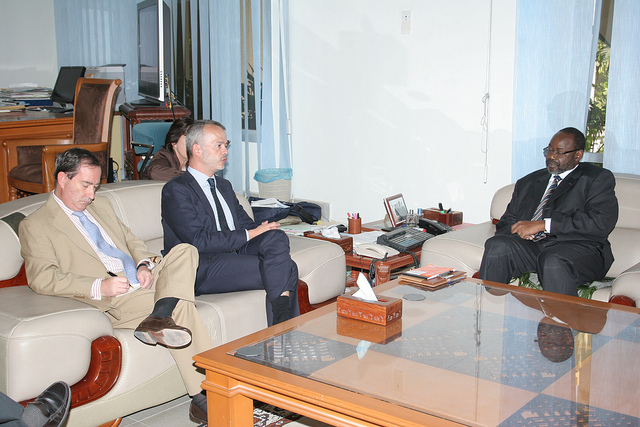}
  &\includegraphics[width=0.40\columnwidth, height=0.35\columnwidth]{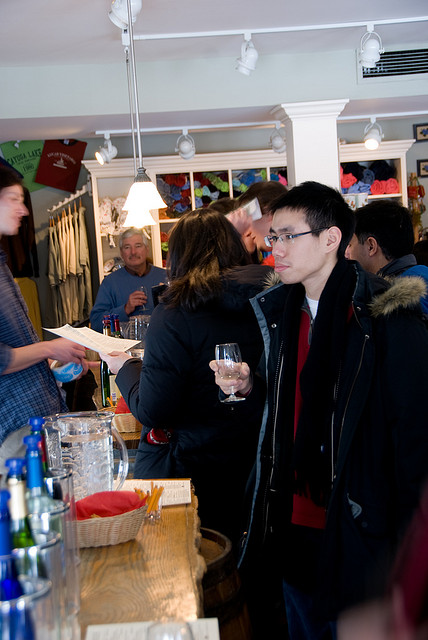}
  &\includegraphics[width=0.40\columnwidth, height=0.35\columnwidth]{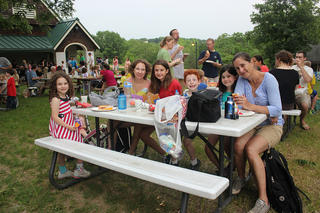}
  &\includegraphics[width=0.40\columnwidth, height=0.35\columnwidth]{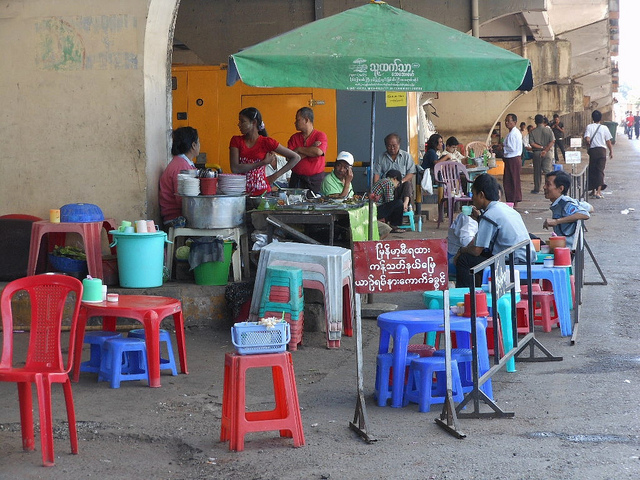}
  &\includegraphics[width=0.40\columnwidth, height=0.35\columnwidth]{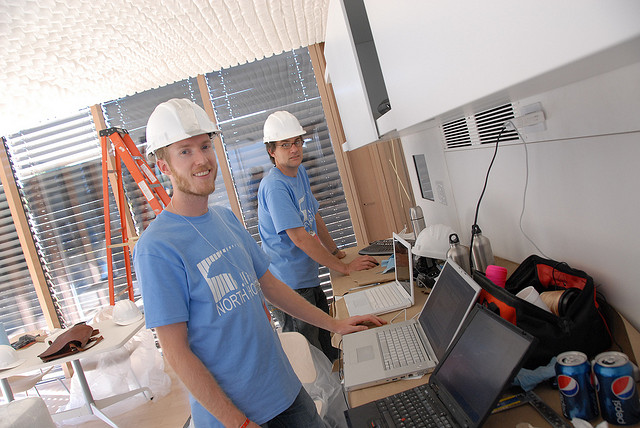}
  \\
  \multirow{1}{*}{\parbox[c][][c]{5mm}{Adv-samp}} & a group of men sitting around a meeting room & a group of people sitting at a bar drinking wine& a group of friends enjoying lunch outdoors at a outdoor event& a group of people sitting at tables outside & a couple of men that are working on laptops\\
  \\
  \multirow{1}{*}{\parbox[c][][c]{5mm}{Base-samp}} & \multicolumn{5}{c}{$\cdots\cdots\cdots\cdots\cdots$\quad\quad a group of people sitting around a table \quad\quad$\cdots\cdots\cdots\cdots\cdots$ }\\

  &\includegraphics[width=0.40\columnwidth, height=0.35\columnwidth]{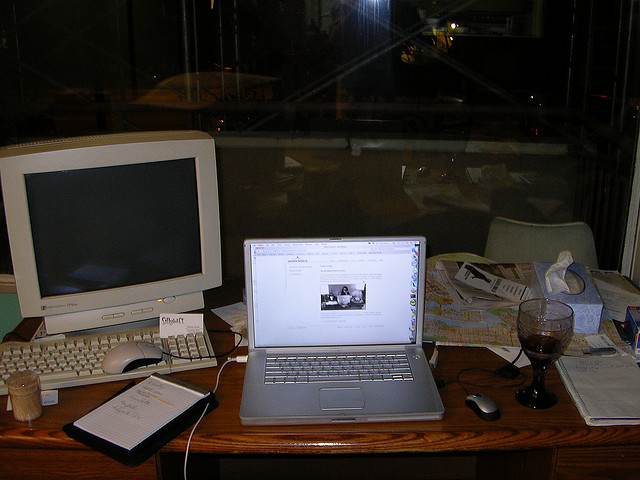}
  &\includegraphics[width=0.40\columnwidth, height=0.35\columnwidth]{images/COCO_val2014_000000388481.jpg}
  &\includegraphics[width=0.40\columnwidth, height=0.35\columnwidth]{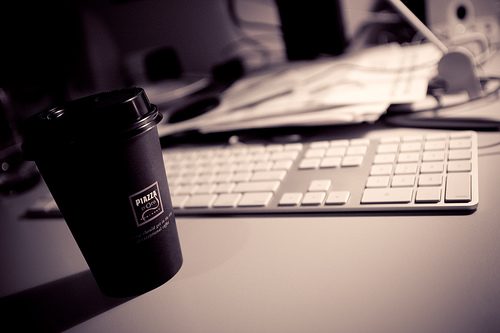}
  &\includegraphics[width=0.40\columnwidth, height=0.35\columnwidth]{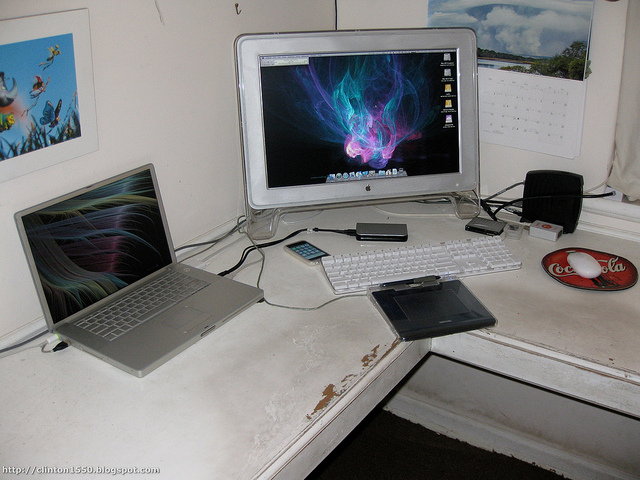}
  &\includegraphics[width=0.40\columnwidth, height=0.35\columnwidth]{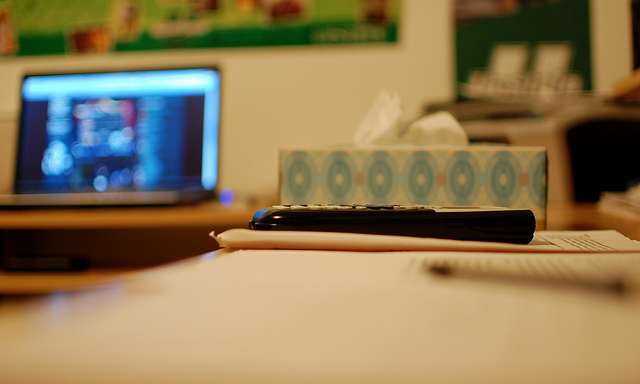}
  \\
  \multirow{1}{*}{\parbox[c][][c]{5mm}{Adv-samp}} & a laptop and a desktop computer sit on a desk& a person is working on a computer screen& a cup of coffee sitting next to a laptop & a laptop computer sitting on top of a desk next to a desktop computer& a picture of a computer on a desk\\
  \\
  \multirow{1}{*}{\parbox[c][][c]{5mm}{Base-samp}} & \multicolumn{5}{c}{$\cdots\cdots\cdots\cdots\cdots$\quad\quad a laptop computer sitting on top of a desk \quad\quad$\cdots\cdots\cdots\cdots\cdots$ }\\
\end{tabular}
\caption{Illustrating diversity across images}
\label{fig:appx4}
\end{center}
\end{figure*}

Adversarial model produces diverse captions across different images containing similar content, whereas the baseline model tends to use common generic captions which is mostly correct but not descriptive. We quantify this by looking at the most frequently generated captions by the baseline model on the test set in \Tableref{tab:repeatbase}. Note that here we consider only the most likely caption according to the model. \Tableref{tab:repeatbase} shows that the baseline model tends to repeatedly generate identical captions for different images. Compared to this, adversarial model is less prone to repeating generic captions, as seen in \Tableref{tab:repeatadv}. This is visualized in Figures~\ref{fig:appx5}, \ref{fig:appx6} and \ref{fig:appx4}. Here we show sets of five images for which the baseline model generates identical generic caption. The five images are picked from among the images corresponding to captions in \Tableref{tab:repeatbase}, starting from the most frequent caption. Some entries are skipped, for example the caption in the last row, to avoid repeated concepts. While the baseline model produces a fixed generic caption for these images, we see that the adversarial model produces diverse captions and which is often more specific to the contents of image.  For example, in the last row of \Figref{fig:appx5} we can see that the baseline model produces the generic caption "\emph{a man riding skis down a snow covered slope}", whereas captions produced by the adversarial model include more image specific terms like "\emph{jumping}", "\emph{turn}", "\emph{steep}" and "\emph{corss country skiing}".

\begin{table}[tb]
\centering
\small
\begin{tabular}{p{5.0cm}p{1.2cm}p{1.2cm}}
\toprule
\bf Sentence & \bf \# baseline &\bf \# adversarial\\              
 \cmidrule(lr){1-1}\cmidrule(lr){2-2}\cmidrule(lr){3-3}
 a man riding a wave on top of a surfboard & 54 & 20 \\
 a bathroom with a toilet and a sink & 44 & 1 \\
 a baseball player swinging a bat at a ball & 37 & 2\\
 a man riding skis down a snow covered slope & 29 & 5 \\
 a man holding a tennis racquet on a tennis court & 26 & 3 \\
 a bathroom with a sink and a mirror& 25 & 3\\
 a man riding a snowboard down a snow covered slope& 24 & 4\\
 a baseball player holding a bat on a field& 22 & 1\\
 a man riding a skateboard down a street& 21 & 2\\
 a bus is parked on the side of the road& 20 & 0\\
 $\cdots$& $\cdots$  & $\cdots$ \\
 \bottomrule
\end{tabular}
\caption{Most frequently repeated captions generated by the baseline model on the test set of 5000 images. The caption "a man riding a wave on top of a surfboard" is also the most frequently generated caption by the adversarial model, albeit less than half the times of the baseline model.}
\label{tab:repeatbase}
\end{table}

\begin{table}[tb]
\centering
\small
\begin{tabular}{p{5.0cm}p{1.2cm}p{1.2cm}}
\toprule
\bf Sentence &\bf \# adversarial& \bf \# baseline \\              
 \cmidrule(lr){1-1}\cmidrule(lr){2-2}\cmidrule(lr){3-3}
 a man riding a wave on top of a surfboard & 20 & 54 \\
 a skateboarder is attempting to do a trick & 16 & 0 \\
 a female tennis player in action on the court & 16 & 0\\
 a living room filled with furniture and a flat screen tv & 15 & 0 \\
 a bus that is sitting in the street & 15 & 3 \\
 a long long train on a steel track& 11 & 0\\
 a close up of a sandwich on a plate& 10 & 0\\
 a baseball player swinging at a pitched ball& 10 & 0\\
 a bus that is driving down the street& 9 & 1\\
 a boat that is floating in the water& 8 & 3\\
 $\cdots$& $\cdots$  & $\cdots$ \\
 \bottomrule
\end{tabular}
\caption{Most frequently repeated captions generated by the adversarial model on the test set of 5000 images. }
\label{tab:repeatadv}
\end{table}

\end{document}